\documentclass{elsart}
\usepackage{amsmath}
\usepackage{comment}
\usepackage{epsfig}
\usepackage{graphicx}
\usepackage{epsfig}
\usepackage{amssymb}
\usepackage{psfig}
\usepackage[]{subfigure}
\begin{document}
\begin{frontmatter}
\title{Image compression   by rectangular wavelet transform}
\author[slava]{Vyacheslav Zavadsky}
\ead{vyacheslavz@semiconductor.com}  
\address[slava]{Semiconductor Insights, R\&D department, 3000 Solandt Road, Kanata, ON, Canada. K2K2X2}
 
\begin{abstract}
 We study  image compression  by a separable wavelet  basis 
 $\big\{\psi(2^{k_1}x-i)\psi(2^{k_2}y-j),$ 
 $\phi(x-i)\psi(2^{k_2}y-j),$
 $\psi(2^{k_1}(x-i)\phi(y-j),$
 $\phi(x-i)\phi(y-i)\big\},$ 
 where  $k_1, k_2 \in \mathbb{Z}_+$; 
 $i,j\in\mathbb{Z}$; and   $\phi,\psi$ are elements of a standard biorthogonal wavelet basis in $L_2(\mathbb{R})$.
 Because $k_1\ne k_2$, the supports of the basis elements are rectangles, and the corresponding transform is known as the {\em rectangular wavelet transform}.
 We prove that if one-dimensional  wavelet basis has $M$ dual vanishing moments then the rate of approximation by $N$ coefficients of rectangular wavelet transform  is   $\mathcal{O}(N^{-M}\log^C N)$ for  functions with  mixed derivative of order $M$ in each direction.
 
The square wavelet transform yields  the approximation rate is $\mathcal{O}(N^{-M/2})$  for  functions with all derivatives of the total order $M$.  Thus, the rectangular wavelet transform can outperform the square one if an image has a mixed derivative.
We provide  experimental comparison  of image  compression  which shows that rectangular wavelet transform outperform the square one. 
 
\end{abstract}
\end{frontmatter}
\section{Introduction}
\label{s:intr}

Wavelet analysis has become a powerful tool for image and signal processing. 
Initially developed for approximations and analysis of functions of a single real variable, wavelet techniques was almost immediately generalized for the case of two and many variables.

We will start our discussion with the most recent generalization, {\em non-separable wavelets}. 
In this construction, lattice of integers in one dimensional case is replaced by the quincunx or a general lattice in the multi--dimensional case, and a wavelet analysis  is derived directly for this lattice (see 
 \cite{KovacevicS:97b,uytterhoeven97redblack}).

Although the non-separable approach  now attracts the majority of researchers, we will focus this paper on comparison of the {\em square} and the {\em rectangular} separable wavelets for the  two-dimensional case. 
We will show that the rectangular wavelet transform in some applications  is more suitable than classical square one. 
The ``rectangularization'' technique can  also be  applicable to  some non-separable wavelet constructions.

Separable approaches involve building a basis in $L_2(\mathbb{R}^d)$ using elements of a wavelet  basis for $L_2(\mathbb{R})$.
DeVore {\em et al.} in  \cite{Devore92} and numerous other researchers studied the  following basis in $L_2(\mathbb{R}^2)$:
\begin{multline}
\big\{
\phi(2^lx-i)\phi(2^ly-j), \psi(2^kx-i)\psi(2^ky-j), \\ \phi(2^kx-i)\psi(2^ky-j),   \psi(2^kx-i)\phi(2^ky-j) 
\big\},
\end{multline}
where $\phi$ is a {\em scaling function}, $\psi$ is the corresponding wavelet, $l\in\mathbb{Z}$, $k\ge l$, $k\in\mathbb{Z}$, and $i,j\in\mathbb{Z}$.

Because the supports of the basis functions are localized in the squares with sizes $\mathcal{O}(2^{-k}\times 2^{-k})$, the decomposition on this basis is called the {\em square wavelet transform}. 
It is well known that if a function has bounded (in a certain space) derivatives of the total order $M$ and $\tilde\psi$ has $M$ vanishing moments then the rate of approximation by $N$ elements  of this basis is $\mathcal{O}(N^{-M/2})$.

\begin{figure}[tbhp]
\centering
\subfigure[] {
	\includegraphics[width=5cm]{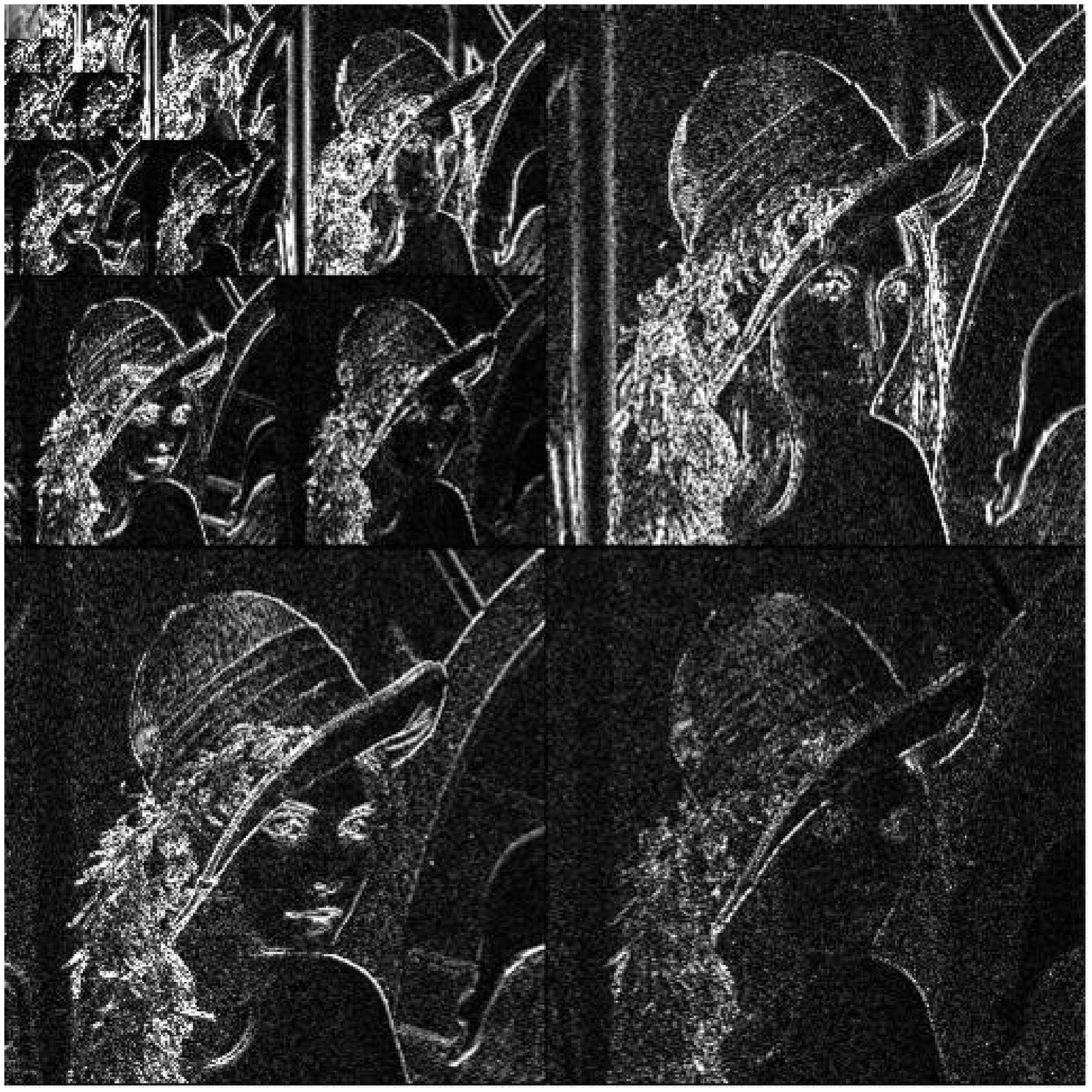}
	\label{lena-haar-square}
	}
\subfigure[] {
	\includegraphics[width=5cm]{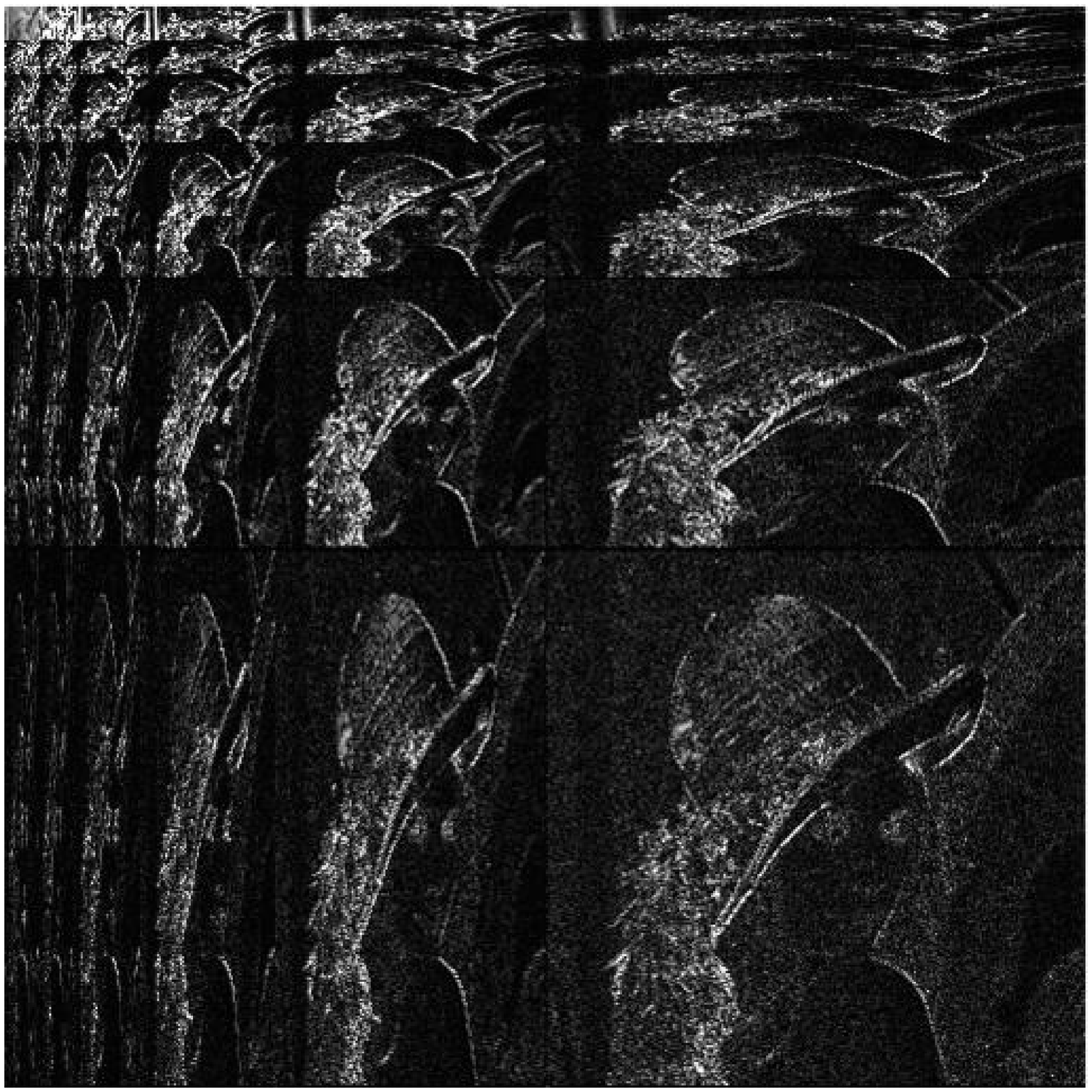}
	\label{lena-haar-rect}
	}

\caption{The Lena images decomposed by square and rectangular wavelet transforms based on the Haar basis.}
\label{lena-haar} 
\end{figure}

{\em Rectangular wavelet transform} is inspired by approximation theory technique known as ``sparse grid'', or ``hyperbolic cross'' (see, for example, \cite{temlyakov:89a}).
The square wavelet transform (and it's higher dimensional analogs)  suffers from so called ``curse of dimensionality''. 
In order to get the same precision of approximation as  by $N$ coefficients  in the one dimensional case, one needs to utilize $\mathcal{O}(N^d)$ coefficients for the function of $d$ variables.
The  ``sparse grid'' technique allows to build a basis for functions of many variables  that yields the same  rate of convergence (up to a logarithmic factor)   as in the case of one variable.
In application to the construction of a   wavelet basis, it yields a basis
\begin{multline}
\big\{\psi(2^{k_1}x-i)\psi(2^{k_2}y-j), \-
 \phi(2^lx-i)\psi(2^{k_2}y-j), \\
 \psi(2^{k_1}(x-i)\phi(2^ly-j,\phi(x-i)\phi(y-i))\big\},
 \end{multline}
where  $k_1, k_2\ge l$, $k_1, k_2 \in \mathbb{Z}$, 
 $i,j\in\mathbb{Z}$ and $l$ is a fixed integer number.  
 Because elements of the basis are supported on rectangles of size 
 $\mathcal{O}(2^{-k_1}\times 2^{-k_2})$, and $k_1\ne k_2$, the corresponding transform can be called {\em  rectangular}.

The application of this basis to the problem of multivariate surface denoising was studied  in \cite{neumann-multivariate} and \cite{VLZ:denois}. The approximation properties of the rectangular wavelet transform in Besov spaces was studied DeVore {\it et al.} in  \cite{devore:96}, and by Zavadsky  in \cite{VLZ:mixed}.

This paper aims at providing a self contained study of the approximation properties of the rectangular wavelet transform from the perspective of image processing and benchmarking of the  rectangular transforms with respect to the square one for image compression.

 We  show  that if the approximated function has the mixed derivative of total order $2M$ and $\tilde\psi$ has $M$ vanishing moments, then the rate of approximation by $N$ elements of this basis is $\mathcal{O}(N^{-M}\log^C N)$.
 
The rest of the paper is organized as follows. In section \ref{sec:oned}, we recall some basic properties of one dimensional wavelet bases. 
In section \ref{sec:motivation}, we consider our motivations for migration from square wavelet transform to rectangular one. 
In section \ref{sec:approximation}, we consider approximation rates of the  non-linear approximations by the rectangular wavelet basis in $L_p(\mathbb{R}^2)$, $1\le p < \infty$. 
In section \ref{sec:experiments}, we provide numerical experiments results  on standard test images.

\section{One-dimensional wavelet transform}
\label{sec:oned}
Biorthogonal bases of compactly supported wavelets were build in \cite{Cohen92biorthogonal}.  
Following this paper, we assume that we have four compactly supported   functions $\phi,\tilde\phi,\psi,\tilde\psi\in L_\infty(\mathbb{R})$  and coefficients vectors $h,\tilde h, g, \tilde g$ that satisfy the following conditions:
\begin{align}
\phi(x)&=\sum_{i\in\mathbb{Z}}h_i\phi(2x-i),\label{phi_exp}\\
\tilde\phi(x)&=\sum_{i\in\mathbb{Z}}\tilde h_i\tilde\phi(2x-i),\\
\psi(x)&=\sum_{i\in\mathbb{Z}}g_i\phi(2x-i),\label{psi_exp}\\
\tilde\psi(x)&=\sum_{i\in\mathbb{Z}}\tilde g_i\tilde\phi(2x-i),\\
<\phi(x),\tilde\phi(x-i)>&=\begin{cases} 1 \text{ if $i=0$ },\\ 0\text { otherwise, }\end{cases}i\in\mathbb{Z},\label{biortheq2}\\
<\psi,\tilde\psi(2^kx-i)>&=\begin{cases} 1 \text{ if $i=0$ and $k=0$ },\\ 0\text { otherwise, }\end{cases} i,k\in\mathbb{Z},\\
<\phi,\tilde\psi(2^kx-i)>&=\begin{cases} 1 \text{ if $i=0$ and $k=0$ },\\ 0\text { otherwise, }\end{cases} i\in\mathbb{Z},k\in\mathbb{Z}_+,\\
<\psi,\tilde\phi(2^kx-i)>&=\begin{cases} 1 \text{ if $i=0$ and $k=0$ },\\ 0\text { otherwise, }\end{cases} i\in\mathbb{Z},k\in\mathbb{Z}_-,
\label{biortheq1}
\end{align}
where $<f_1(x),f_2(x)>=\int f_1(x) f_2(x)\,dx$.
We further assume that the basis has {\em $M$ dual vanishing moments}:
\begin{equation}
\int x^k\tilde\psi(x)\, dx=0,\ k=0,\dots, M-1.
\end{equation}

It can be shown that for any scale $l\in\mathbb{Z}$ any function $f\in L_2(\mathbb{R})$ can be represented as
\begin{align}
f(x)=&\sum_{i\in\mathbb{Z}}2^l\phi(2^lx-i)\alpha_{l,i}+\sum_{k=l}^{\infty}2^k\psi(2^kx-i)\beta_{k,i},\label{1d:expansion}\\
\text{ where } \alpha_{k,i}=&<f, \tilde\phi(2^kx-i)>,\\
							\beta_{k,i}=&<f, \tilde\psi(2^kx-i)>.
\end{align}
The {\em fast wavelet transform\/} allows to quickly go in representation \eqref{1d:expansion} from scale $l$ to $l-1$ and back ({\em inverse fast wavelet transform}) by observing that
\begin{align}
\alpha_{l-1,i}&=<f,\tilde\phi(2^{l-1}x-i)>=\sum_{j\in\mathbb{Z}}\tilde h_j<f,\tilde\phi(2^lx-2i-j)>=\sum_{j\in\mathbb{Z}}\tilde h_j\alpha_{l,2i+j},\label{up:alpha}\\
\beta_{l-1,i}&=<f,\tilde\psi(2^{l-1}x-i)>=\sum_{j\in\mathbb{Z}}\tilde g_j<f,\tilde\phi(2^lx-2i-j)>=\sum_{j\in\mathbb{Z}}\tilde g_j\alpha_{l,2i+j},\label{up:beta}
\end{align}
\begin{multline}
\alpha_{l,i}=<f,\tilde\phi(2^{l}x-i)>=
2^{l-1}\sum_{j\in\mathbb{Z}}\alpha_{l-1,j}<\phi(2^{l-1}x-j),\tilde\phi(2^lx-i)>+\\
+2^{l-1}\sum_{j\in\mathbb{Z}}\beta_{l-1,j}<\psi(2^{l-1}x-j),\tilde\phi(2^lx-i)>+\\
+\sum_{k=l}^{\infty}2^{k-1}\sum_{j\in\mathbb{Z}}\beta_{k-1,j}<\psi(2^kx-j),\tilde\phi(2^lx-i)>.
\end{multline}
According to \eqref{biortheq1} the last term is $0$.
Expanding $\phi(2^{l-1}x-j)$ and $\psi(2^{l-1}x-j)$ by \eqref{phi_exp} and \eqref{psi_exp} and then using \eqref{biortheq2}, one can show that 
\begin{equation}
\alpha_{l,i} = \frac{1}{2}\left [ \sum_{j\in\mathbb{Z}}h_{2j-i}\alpha_{l-1,j}+\sum_{j\in\mathbb{Z}}g_{2j-i}\beta_{l-1,j}\right ]
\label{down}
\end{equation}

In practice, if we process a discrete signal sampled with step $2^{-l}$, it is typically assumed that in expansion \eqref{1d:expansion} $\beta_{k_i}=0$ and $\alpha_{l,i}$ are the discrete values of observed signal.

The wavelet transform is typically used to compress or denoise signals by moving to representation \eqref{1d:expansion} with smaller $l$ by iterative applications of  \eqref{up:alpha},\eqref{up:beta}; setting certain $\beta_{k,i}$ to $0$ (that will reduce data size and filter out random noise); and moving back to the original $l$ by iterative application of  \eqref{down}.
 
\section{Motivation}
\label{sec:motivation}
\begin{figure}[tbhp]
\centering
\subfigure[Scaling function $\phi=\tilde\phi$] {
	\includegraphics[width=5cm]{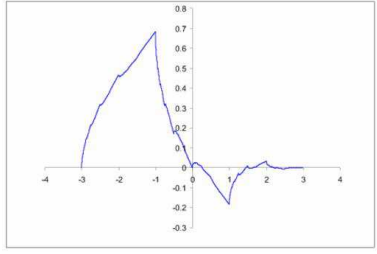}
	\label{wd2-phi}
	}
\subfigure[Wavelet function $\psi=\tilde\psi$] {
	\includegraphics[width=5cm]{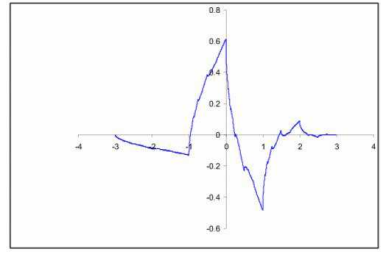}
	\label{wd2-psi}
	}
\subfigure[$\phi(x)\psi(y)$] {
	\includegraphics[width=5cm, height=5cm]{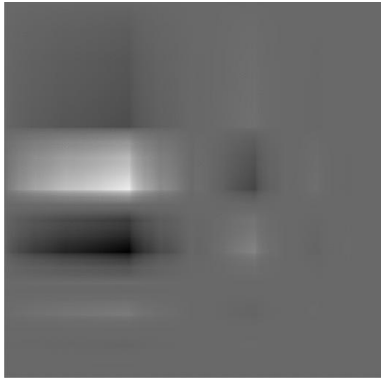}
	\label{wd2-phi-psi}
	}
\subfigure[$\psi(x)\phi(y)$] {
	\includegraphics[width=5cm, height=5cm]{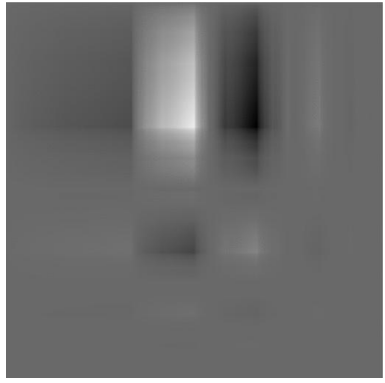}
	\label{wd2-psi-phi}
	}
\subfigure[$\psi(x)\psi(y)$] {
	\includegraphics[width=5cm, height=5cm]{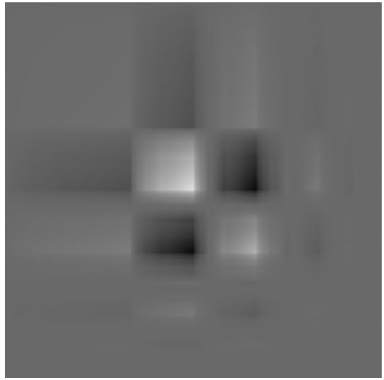}
	\label{wd2-psi-psi}
	}
\caption{D4: the Daubechies orthogonal wavelet with 2 vanishing moments and elements of square wavelet basis built from it.}
\label{dw2-graph} 
\end{figure}

Let us start from considering square wavelet transform build from the D4 Daubechies orthogonal wavelet with 2 vanishing moments, see  Fig. \ref{dw2-graph}. As we see, the $\phi(x)\psi(y)$ term characterizes horizontal edges, $\psi(x)\phi(y)$--vertical edges. Although some authors suggested that $\psi(x)\psi(y)$ can be used to characterize edges in the diagonal directions, we can see that this term have different structure from the previous two and we can expect that the corresponding coefficients will decline with different speed.

\begin{figure}[tbhp]
\centering
\subfigure[D4 decomposition] {
	\includegraphics[width=5cm]{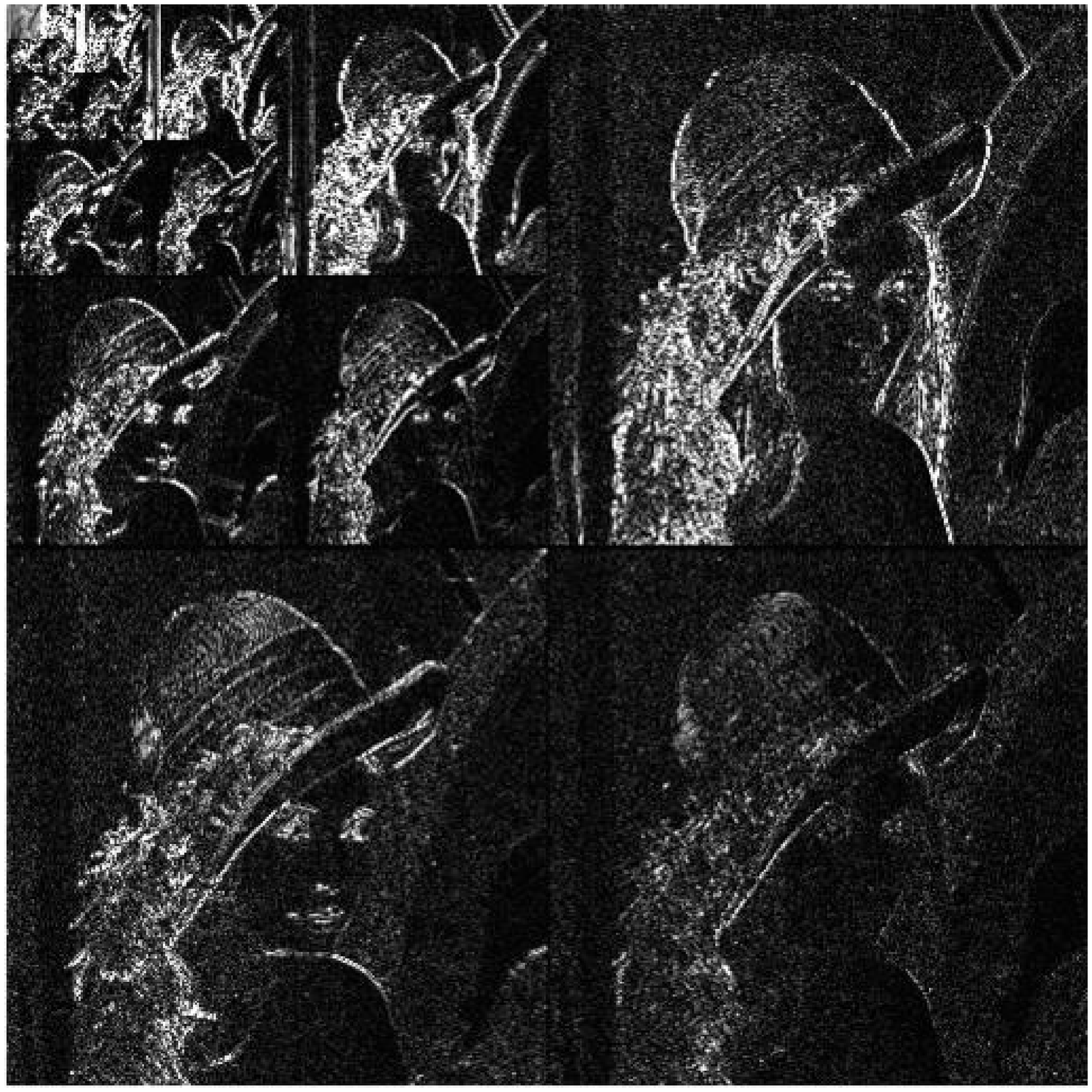}
	\label{lena-square-d4}
}
\subfigure[D4 energy distribution] {
	\includegraphics[width=7cm]{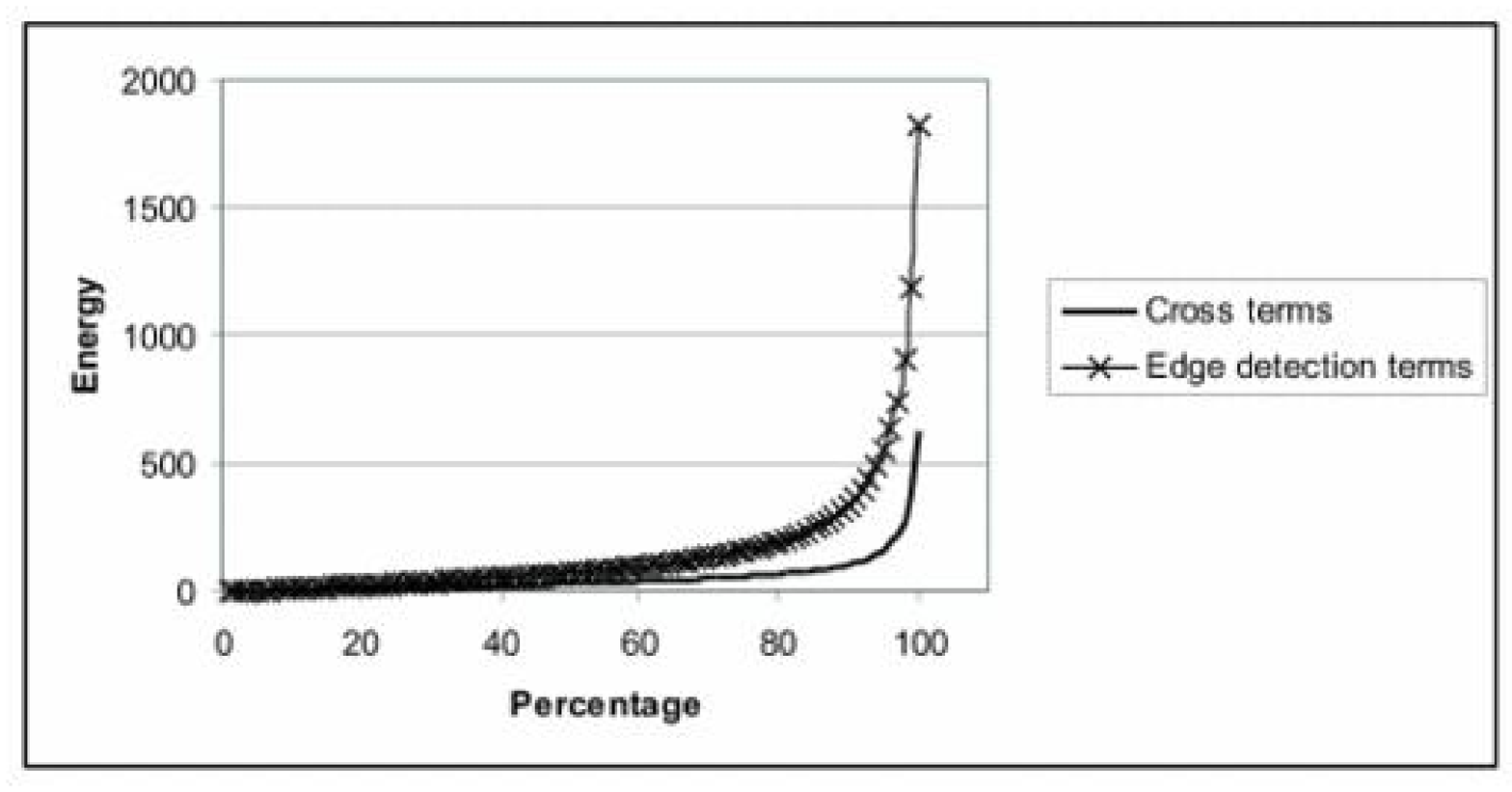}
	\label{lena-square-d4-energy}
}\\
\subfigure[CRF(13,7) decomposition] {
	\includegraphics[width=5cm]{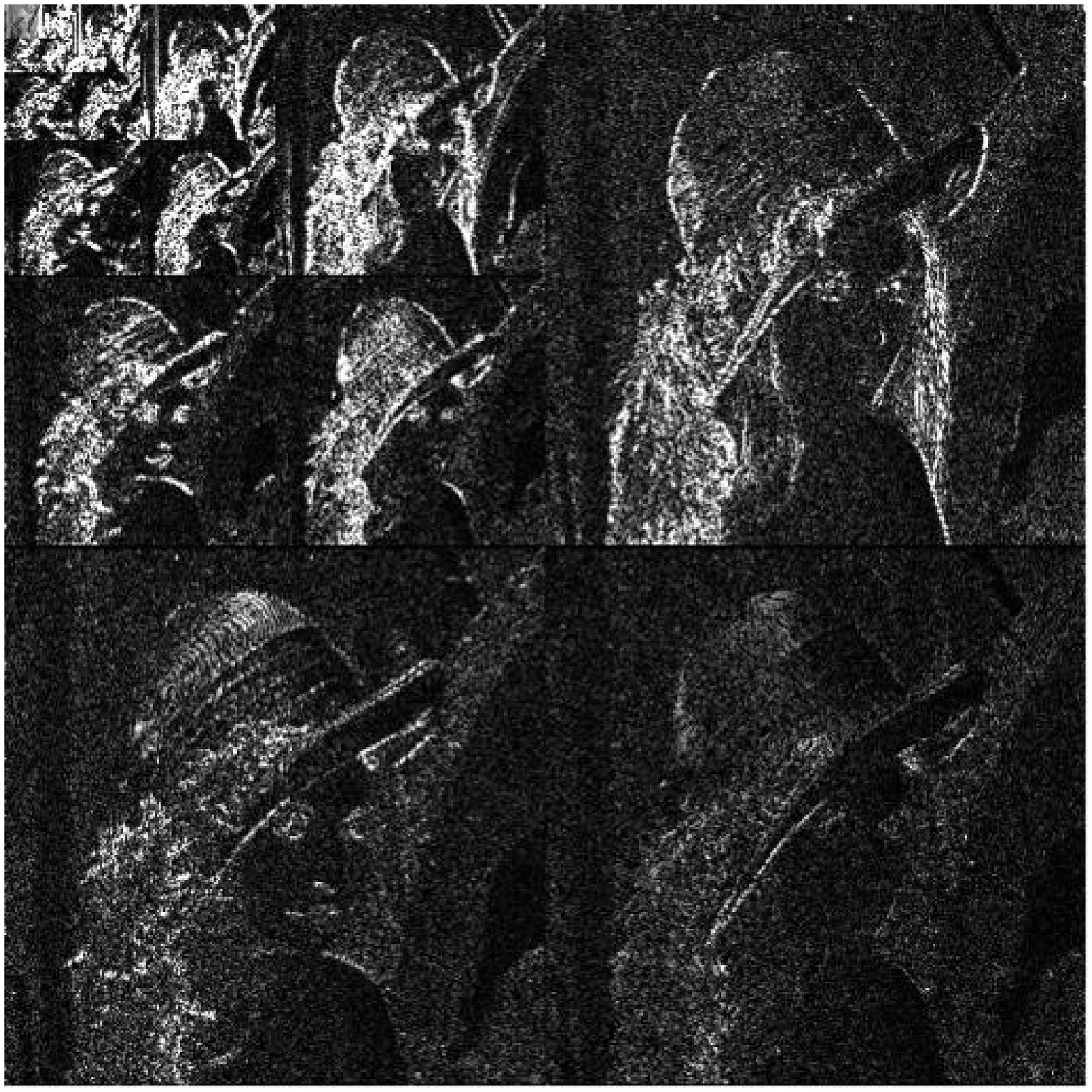}
	\label{lena-crf13-7-square}
}
\subfigure[CRF(13,7) energy distribution] {
	\includegraphics[width=7cm]{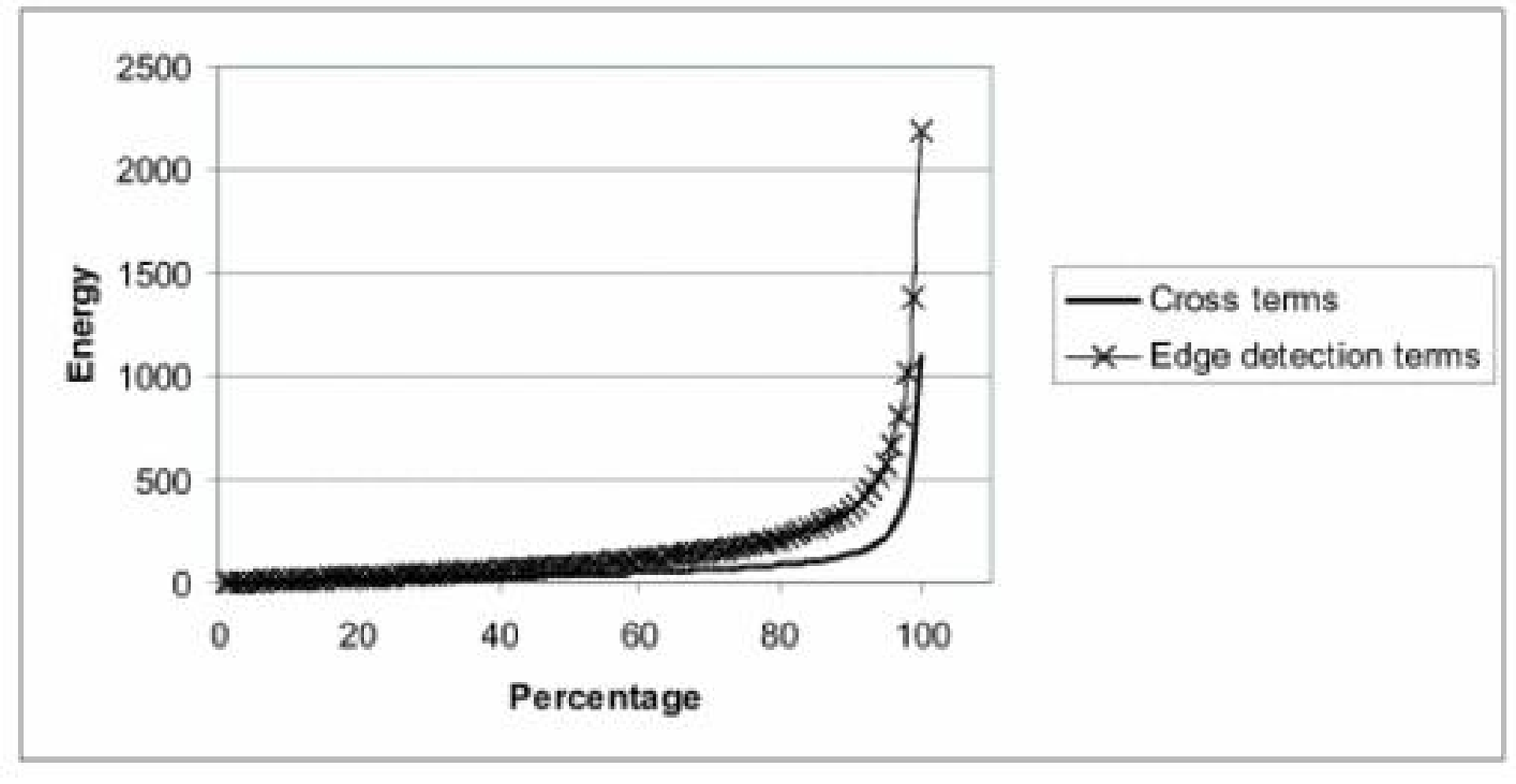}
	\label{lena-crf13-7-square-energy}
}	
\caption{The Lena image decomposed by square wavelet transform built from the Daubechies orthogonal D4 filter and the JPEG2000 CRF(13,7) filter. The energy distribution plots in \ref{sub@lena-square-d4-energy}, \ref{sub@lena-crf13-7-square-energy} show the distribution of  $L_2$ norms of $\beta\phi(x)\psi(y)$, $\beta\psi(x)\phi(y)$ ({\em edge detection terms\/}) and $\beta\psi(x)\psi(y)$ ({\em  cross terms\/}).}
\label{lena-square} 
\end{figure} 
 
Now let us have a look at Figs. \ref{lena-haar-square}, \ref{lena-square-d4}, and  \ref{lena-crf13-7-square}. We can see that 
the diagonal squares, that are shows coefficients of $\psi(2^kx-i)\psi(2^ky-j)$ are  less bright than others. 
 
Now let us consider how the value of coefficients with terms  $\psi(x)\psi(y)$ and $\phi(x)\psi(y)$, $\psi(x)\phi(y)$ can be compared from the theoretical standpoint. The proof of the following lemma is given in Appendix:
\begin{lem}
\label{lemma:combined:bound}
For any wavelet basis with $M$ dual vanishing moments and for any $p\ge 1$ there exist  constants $C,L$ such 
that for any $\sigma_1,\sigma_2>0$ and $\theta_1,\theta_2\in\mathbb{R}$ and for any function $f(x,y)$ with the appropriate  derivative in $L_p^{loc}$:
\begin{multline}
\big | <f(x,y),\tilde\psi(\sigma_1 x-\theta_1)\tilde\phi(\sigma_2 y-\theta_2)>\big|\le \\
\le C \frac{1}{\sigma_1^{M+1-1/p}}\left[
\int_{\frac{\theta_1-L}{\sigma_1}}^{\frac{\theta_1+L}{\sigma_1}}
\int_{\frac{\theta_2-L}{\sigma_2}}^{\frac{\theta_2+L}{\sigma_2}}
 \big|\frac{d^M}{dx^M}f(x,y) \big|^p\, dy\, dx\right]^{1/p},
\end{multline}
\begin{multline}
\big | <f(x,y),\tilde\phi(\sigma_1 x-\theta_1)\tilde\psi(\sigma_2 y-\theta_2)>\big|\le \\
\le C \frac{1}{\sigma_2^{M+1-1/p}}\left[
\int_{\frac{\theta_1-L}{\sigma_1}}^{\frac{\theta_1+L}{\sigma_1}}
\int_{\frac{\theta_2-L}{\sigma_2}}^{\frac{\theta_2+L}{\sigma_2}}
  \big|\frac{d^M}{dy^M}f(x,y) \big|^p\, dy\, dx\right]^{1/p}, 
 \end{multline}
 \begin{multline}
\big | <f(x,y),\tilde\psi(\sigma_1 x-\theta_1)\tilde\psi(\sigma_2 y-\theta_2)>\big|\le \\
\le C \frac{1}{(\sigma_1\sigma_2)^{M+1-1/p}}\left[
\int_{\frac{\theta_1-L}{\sigma_1}}^{\frac{\theta_1+L}{\sigma_1}}
\int_{\frac{\theta_2-L}{\sigma_2}}^{\frac{\theta_2+L}{\sigma_2}}
  \big|\frac{d^{2M}}{dx^Mdy^M}f(x,y) \big|^p\, dy\, dx\right]^{1/p}. 
 \label{one-coef-mix-bound}
 \end{multline}
\end{lem}

\section{Non-linear approximation}
\label{sec:approximation}
\begin{thm} For any wavelet basis with $M$ dual vanishing moments and for any $p\ge\max(1,1/M)$; $q\ge 1$ there exists a constant $C$ such that any function $f(x,y)$ with support localized in $[0,1]^2$ and the mixed derivative $\frac{d^{2M}}{dx^Mdy^M}f(x,y)\in L_p$  for any $N>0$ can be approximated by a function $f_N$ such that
\begin{equation}
\| f-f_N \|_{L_q}\le CN^{-M} \ \big \| \frac{d^{2M}}{dx^Mdy^M}f(x,y)  \big \|_{L_p}\log^C N,
\end{equation}
where $f_N$ has non more than $CN$ non zero coefficients in the expansion on the rectangular wavelet basis.
\end{thm}
\begin{pf}
Denote:
\begin{align}
\mathcal{I} &=\mathbb{Z}^2,\\
\mathcal{L} &=\mathbb{Z}^2_+,\\
\phi_\mathbf{i}&=\phi(x-i_1)\phi(x-i_2),\ (i_1,i_2)\in\mathcal{I},\\
\tilde\phi_\mathbf{i}&=\tilde\phi(x-i_1)\tilde\phi(x-i_2),\\
\psi_{\mathbf{l},\mathbf{i}}&=\psi(2^{l_1}x-i_1)\psi(2^{l_2}y-i_2), \ (l_1,l_2)\in\mathcal{L}, (i_1,i_2)\in\mathcal{I},\\
\tilde\psi_{\mathbf{l},\mathbf{i}}&=\tilde\psi(2^{l_1}x-i_1)\tilde\psi(2^{l_2}y-i_2),\\
\alpha_\mathbf{i}&=<f, \tilde\phi_\mathbf{i}>,\\
\beta_{\mathbf{l},\mathbf{i}}&=<f, \tilde\psi_{\mathbf{l},\mathbf{i}}>,\\
\|\mathbf{l}\|&=l_1+l_2,\ (l_1,l_2)\in\mathcal{L},\\
\mathcal{D}&=\big\| \frac{d^{2M}}{dx^Mdy^M}f(x,y)  \big\|_{L_p}.
\end{align}
Because the set $\big\{\phi_\mathbf{i}, \psi_{\mathbf{l},\mathbf{i}}\big |\mathbf{i}\in\mathcal{I}, \mathbf{l}\in\mathcal{L}\big\}$ is the tensor product of the wavelet basis in $L_q(\mathbb{R})$, it forms a basis in 
$L_q(\mathbb{R}^2)$ and
\begin{equation}
f(x,y)=\sum_{\mathbf{i}\in \mathcal{I}}\alpha_\mathbf{i}\phi_\mathbf{i}+
\sum_{\mathbf{l}\in \mathcal{L}}2^{\|\mathbf{l}\|}\sum_{\mathbf{i}\in \mathcal{I}}\beta_{\mathbf{l},\mathbf{i}}\psi_{\mathbf{l},\mathbf{i}}.
\end{equation}

Because $f(x,y)$ is compactly supported, there exists a constant $C$ such that for any $\mathbf{l}\in\mathcal{L}$
\begin{equation}
\mathcal{N}(\|\mathbf{l}\|)=\big | \big \{ \mathbf{i} \, | \mathbf{i}\in\mathcal{I}, \, \beta_{\mathbf{l},\mathbf{i}}\ne 0 \big \}\big | \le C 2^{\|\mathbf{l}\|}+C,
\label{one-level-n-est}
\end{equation} 
where here and below $C$ denotes the universal constant that is maximum of all appropriate constants in the proof and  depends only on the wavelet basis, $p$, and $q$.

Denote by  $l_0$ the maximum $l$ such that $l^2 \mathcal{N}(l)\le N$.

Define 
\begin{equation}
\epsilon_l = {\mathcal{D}}/{2^{l+\frac{M+1/p}{2}l_0+\frac{M-1/p}{2}l}}.
\end{equation}

Let 
\begin{equation}
f_N=\sum_{\mathbf{i}\in \mathcal{I}}\alpha_\mathbf{i}\phi_\mathbf{i}+
\sum_{\mathbf{l}\in \mathcal{L}}2^{\|\mathbf{l}\|}\sum_{\mathbf{i}\in \mathcal{I}}\tilde\beta_{\mathbf{l},\mathbf{i}}\psi_{\mathbf{l},\mathbf{i}},
\end{equation}
where 
\begin{equation}
\tilde\beta_{\mathbf{l},\mathbf{i}}=
\begin{cases}
0, \text{ if $\| \mathbf{l}\| >l_0$ and $\beta_{\mathbf{l},\mathbf{i}}\le \epsilon_{\|\mathbf{l}\|}$,}\\
\beta_{\mathbf{l},\mathbf{i}} \text{ otherwise}.
\end{cases}
\end{equation}
Now let us calculate the number of non-zero coefficients $\tilde N$ in $\tilde\beta_{\mathbf{l},\mathbf{i}}$.
From  \eqref{one-coef-mix-bound} it follows that for any fixed $\mathbf{l}$
\begin{equation}
\|\beta_{\mathbf{l},\cdot}\|_{l_p}\le C \mathcal{D}/2^{\|\mathbf{l}\|(M+1-1/p)}.
\end{equation}
Let us note that if $\|\beta\|_{l_p}<A$ then the number of coefficients bigger than $\epsilon>0$ 
in $\beta$ can not exceed $A^p/\epsilon^p$. Taking into  account \eqref{one-level-n-est}, we have
\begin{equation}
\tilde N \le C\sum_{l=0}^{l_0} l (2^l+C) +\sum_{l=l_0+1}^{\infty} l \frac{\mathcal{D}^p}{2^{lp(M+1-1/p)}\epsilon_l^p}\le
CN.  
\end{equation}

Because $\psi\in L_\infty$ and is compactly supported,
\begin{multline}
\| f-f_N \|_{L_{q}}\le C \sum_{l=l_0+1}^\infty l2^l\epsilon_l=\\= C \mathcal{D} \sum_{l=l_0+1}^\infty l/2^{\frac{M+1/p}{2}l_0+\frac{M-1/p}{2}l}\le C N^{-M} \big\| \frac{d^{2M}}{dx^Mdy^M}f(x,y)  \big\|_{L_p} \log^C N
\end{multline}
\qed
\end{pf}
\section{Experiments}
\label{sec:experiments}
\begin{figure}[tbhp]
\centering
\subfigure[The original "Mandrill" $512\times\- 512$ image.] {
	\includegraphics[width=4cm]{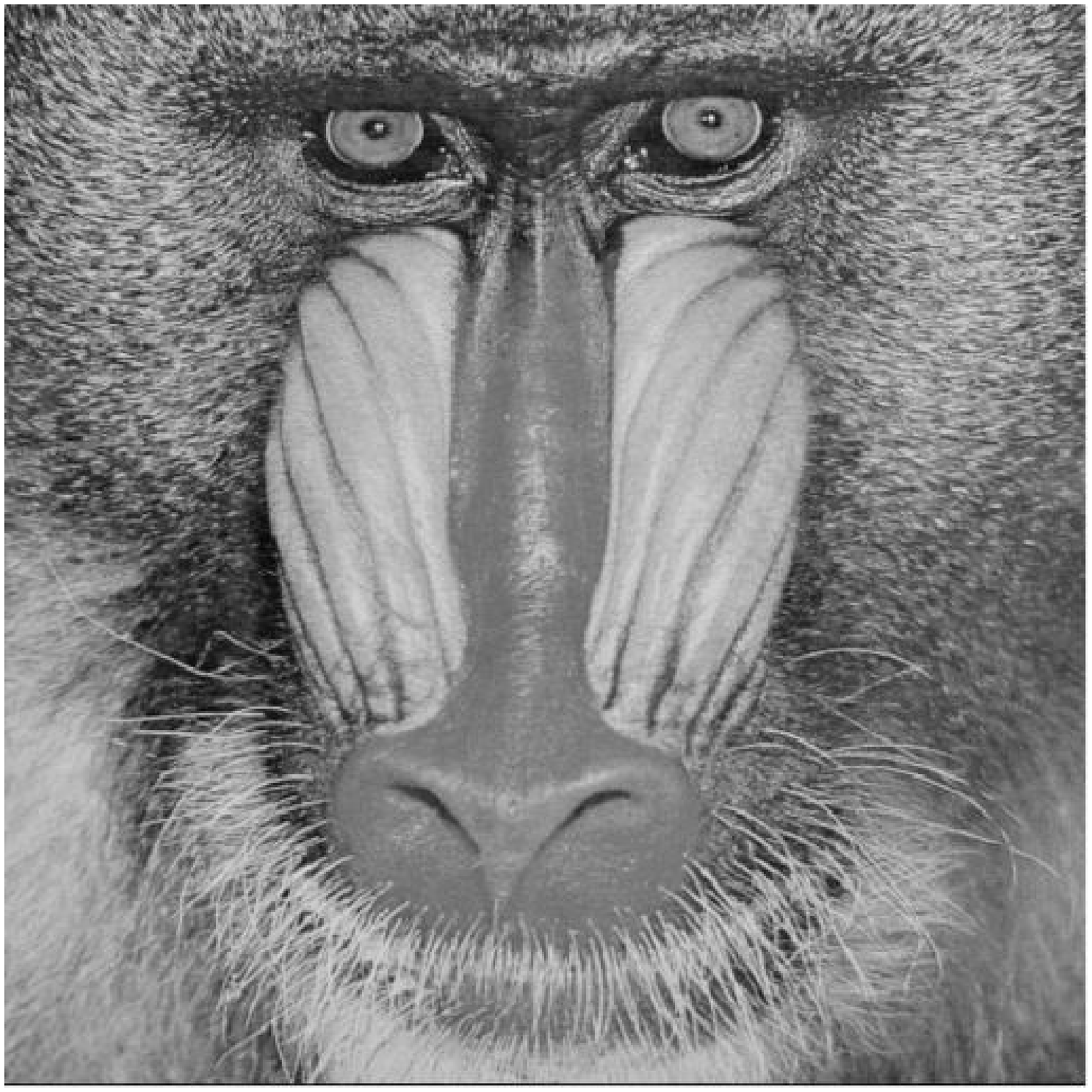}
	\label{exp:mandrill:orig}
}
\subfigure[D4 orthogonal wavelet, compression 1:83, rectangular transform.] {
	\includegraphics[width=4cm]{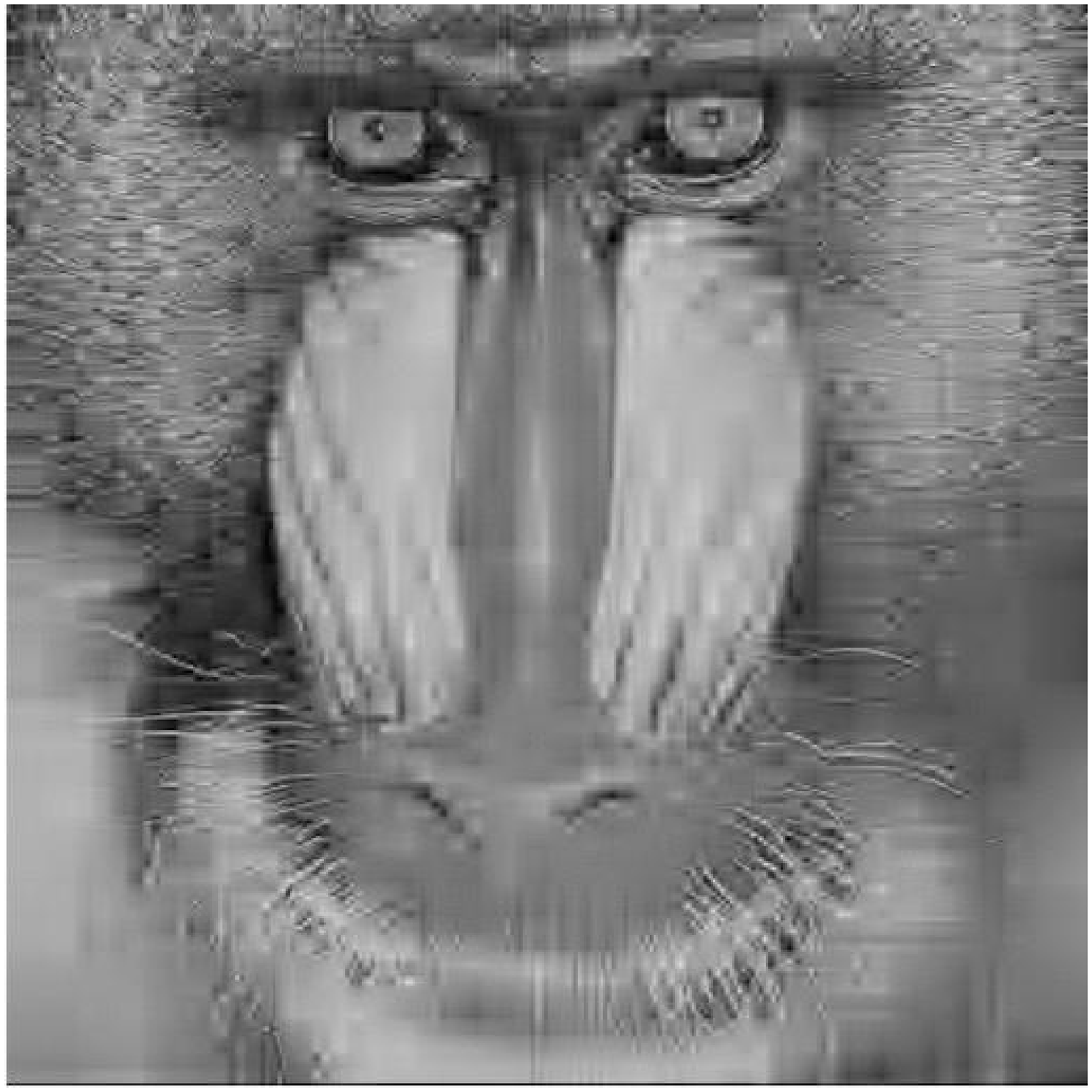}
	\label{exp:mandrill:rect:d2:83}
}
\subfigure[D4 orthogonal wavelet, compression 1:83, square transform.] {
	\includegraphics[width=4cm]{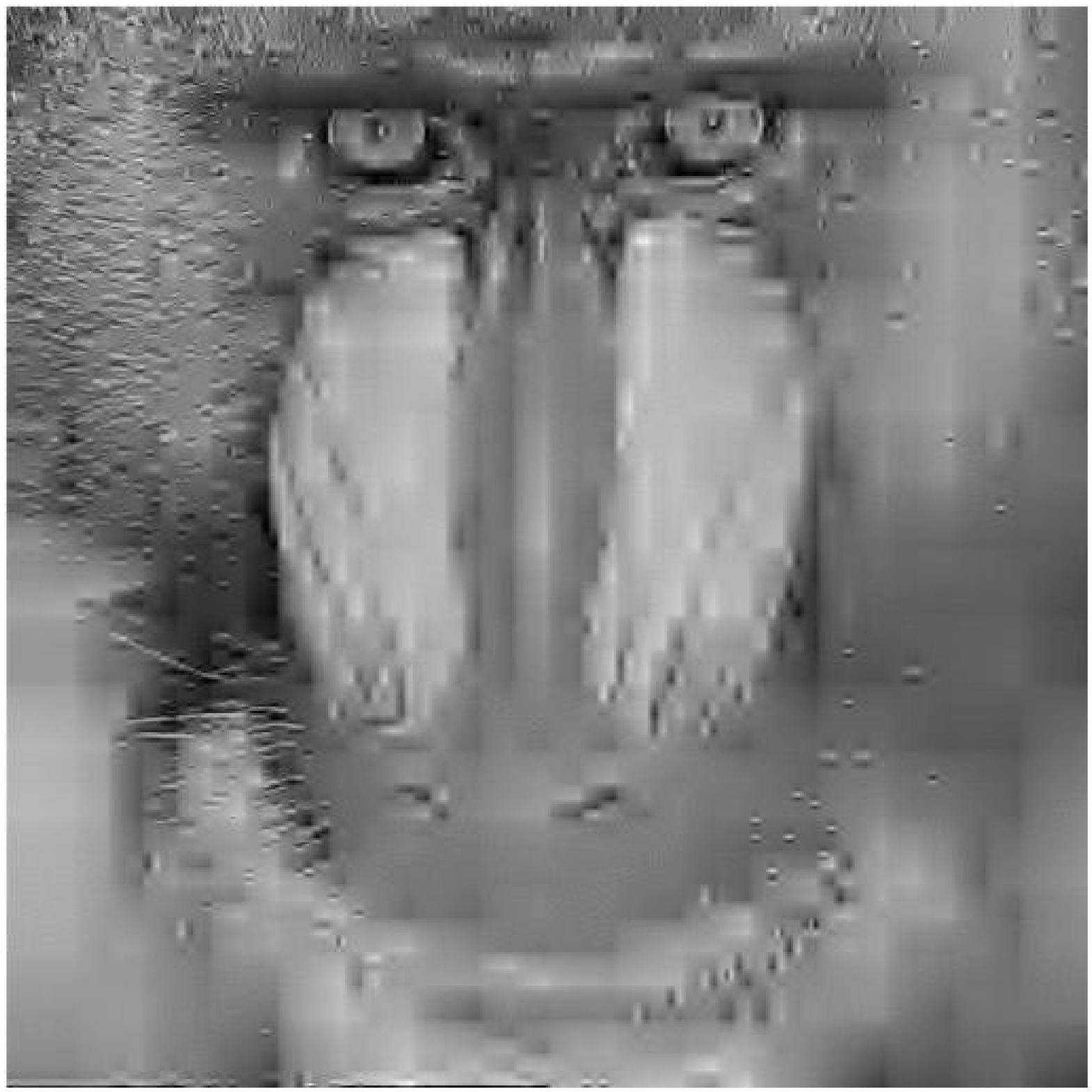}
	\label{exp:mandrill:square:d2:83}
	}
\subfigure[CRF(13,7) biorthogonal wavelet, compression 1:87, rectangular transform.] {
	\includegraphics[width=4cm]{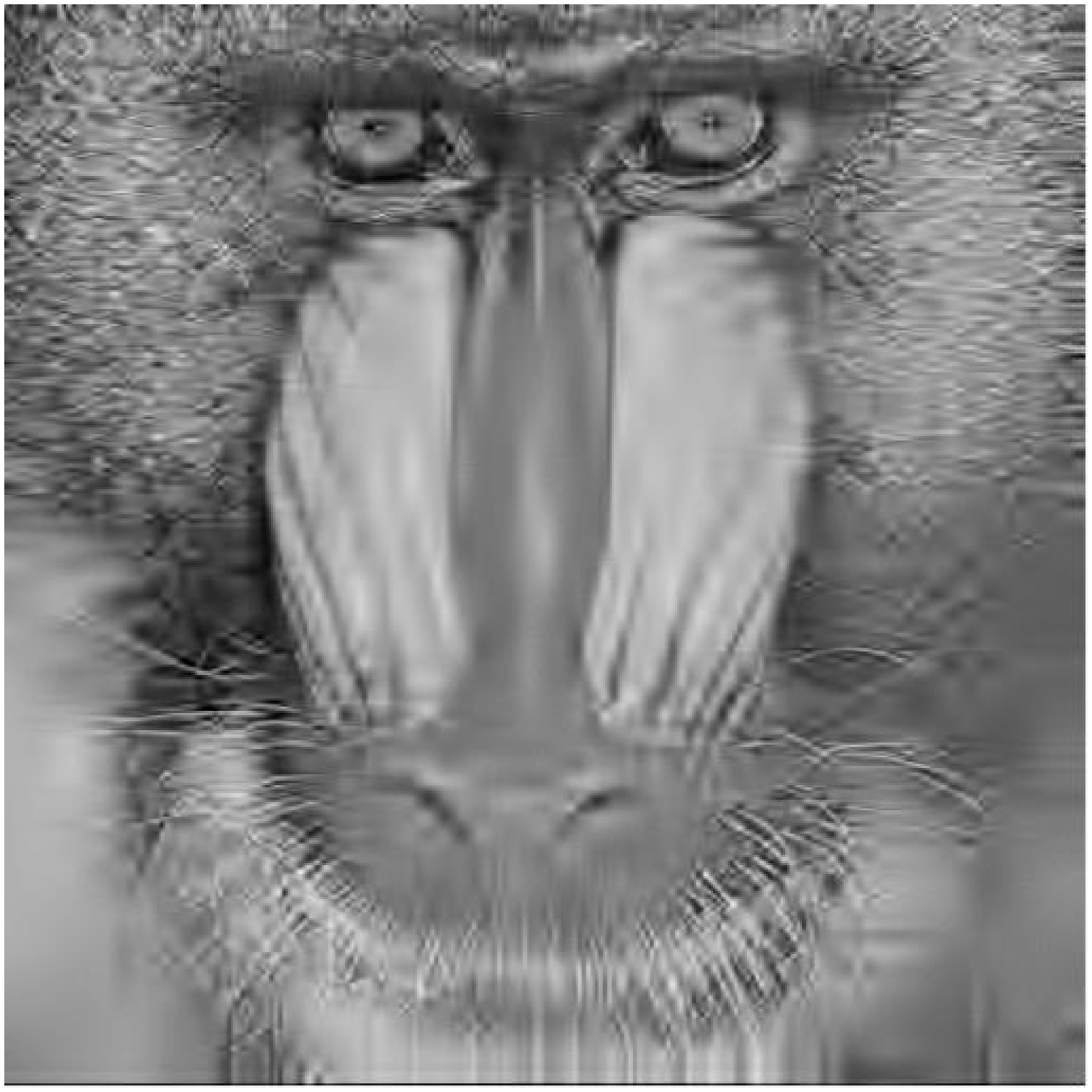}
	\label{exp:mandrill:rect:crf137:87}
	}
\subfigure[CRF(13,7) biorthogonal wavelet, compression 1:81, square transform.] {
	\includegraphics[width=4cm]{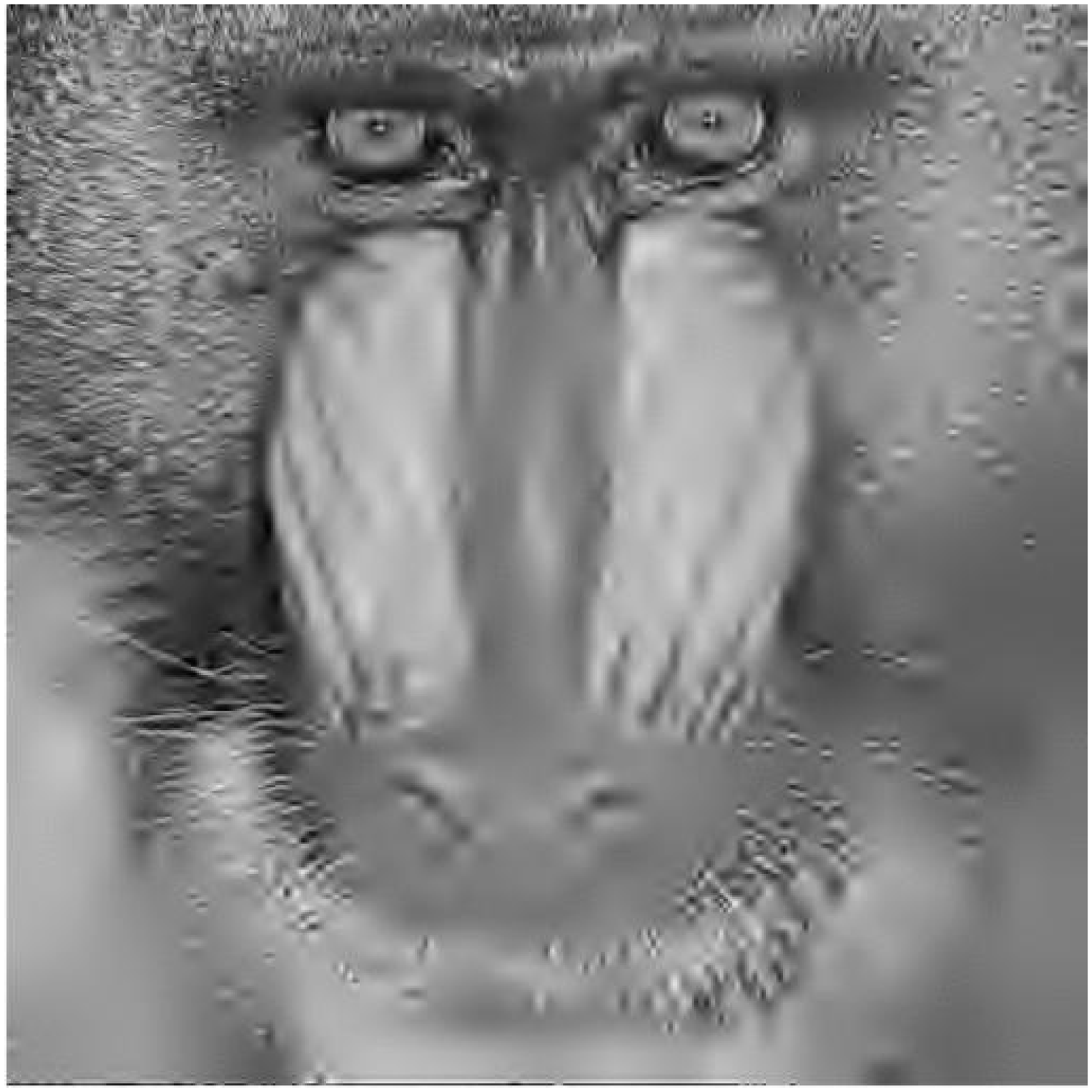}
	\label{exp:mandrill:square:crf137:81}
	}
\subfigure[D4 orthogonal wavelet, compression 1:172, rectangular transform.] {
	\includegraphics[width=4cm]{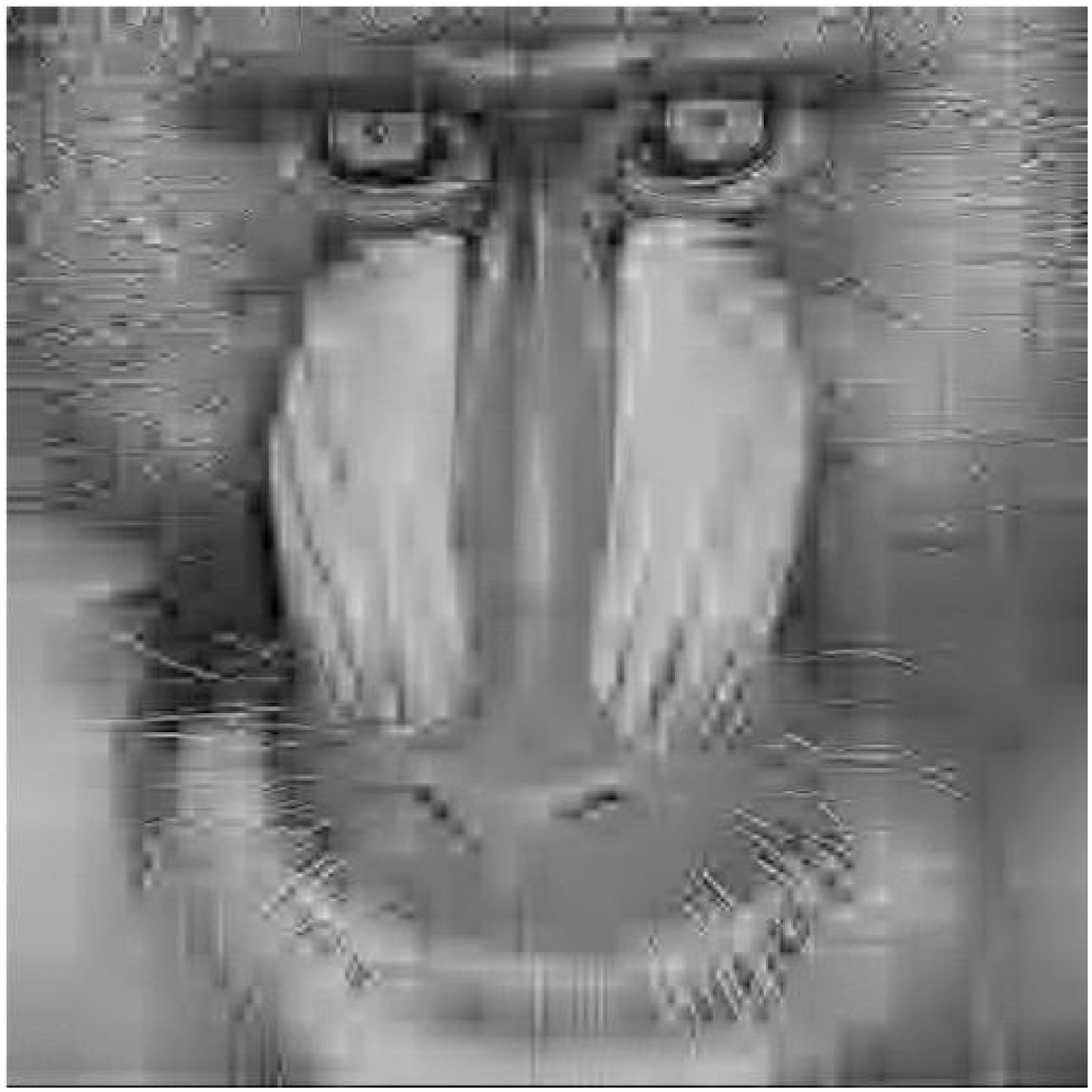}
	\label{exp:mandrill:rect:d2:172}
	}
\subfigure[D4 orthogonal wavelet, compression 1:167, square transform.] {
	\includegraphics[width=4cm]{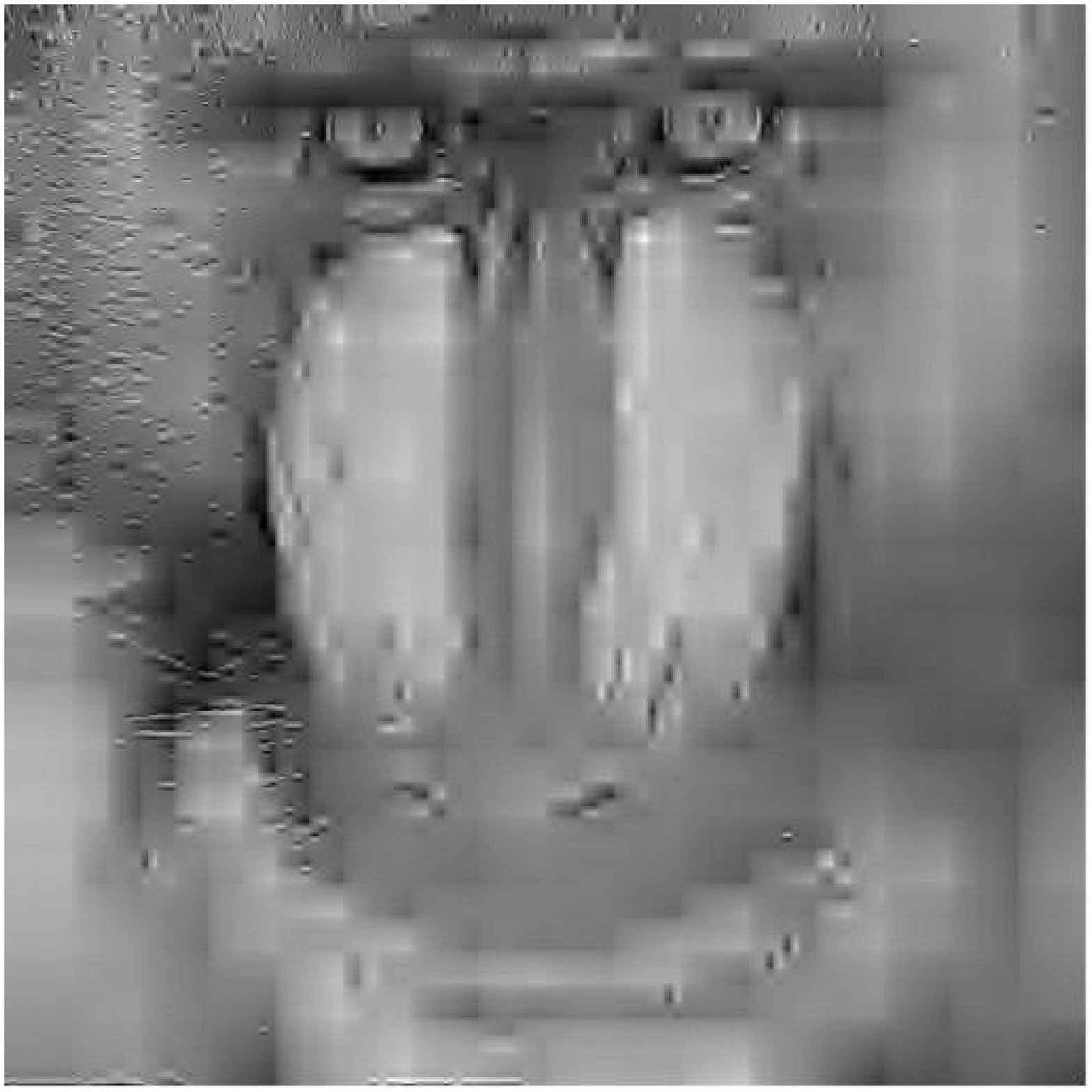}
	\label{exp:mandrill:square:d2:167}
	}
\subfigure[CRF(13,7) biorthogonal wavelet, compression 1:163, rectangular transform.] {
	\includegraphics[width=4cm]{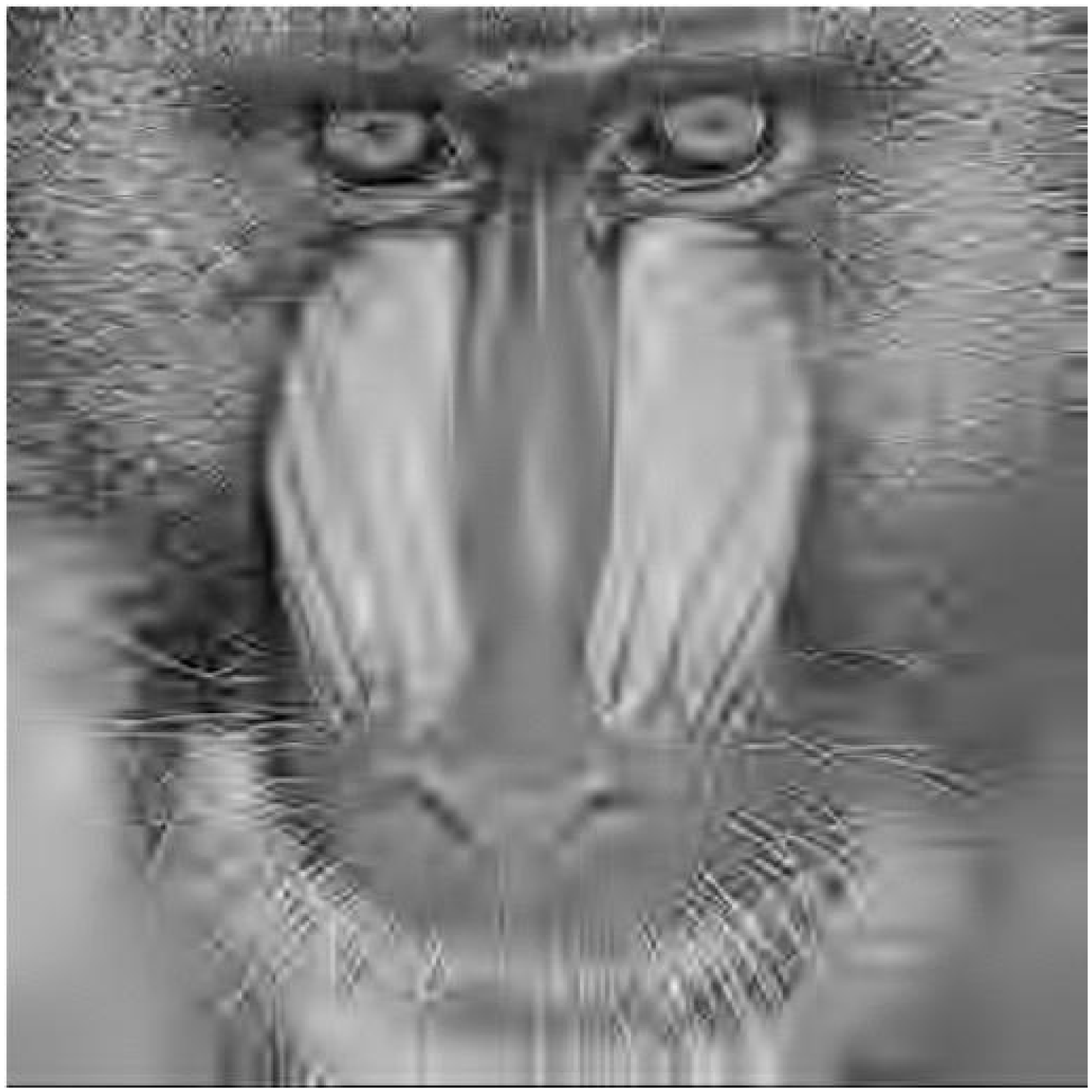}
	\label{exp:mandrill:rect:crf137:163}
	}
\subfigure[CRF(13,7) biorthogonal wavelet, compression 1:166, square transform.] {
	\includegraphics[width=4cm]{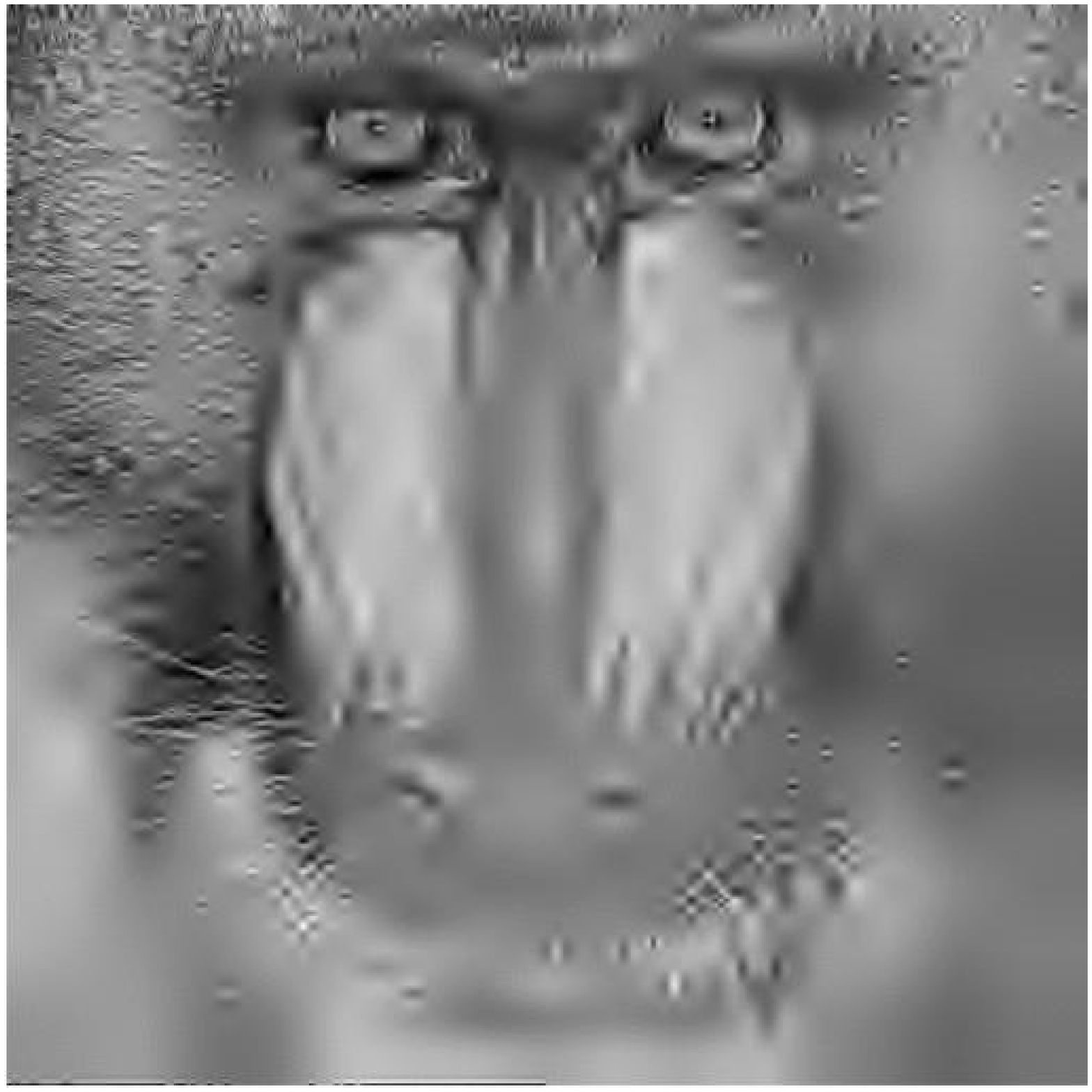}
	\label{exp:mandrill:square:crf137:166}
	}
\caption{The Mandrill  image compressed by the rectangular and square wavelet transforms.}
\label{exp-mandrill} 
\end{figure}
\begin{figure}[tbhp]
\centering
\subfigure[The original "House" $512\times 512$ image.] {
	\includegraphics[width=4cm]{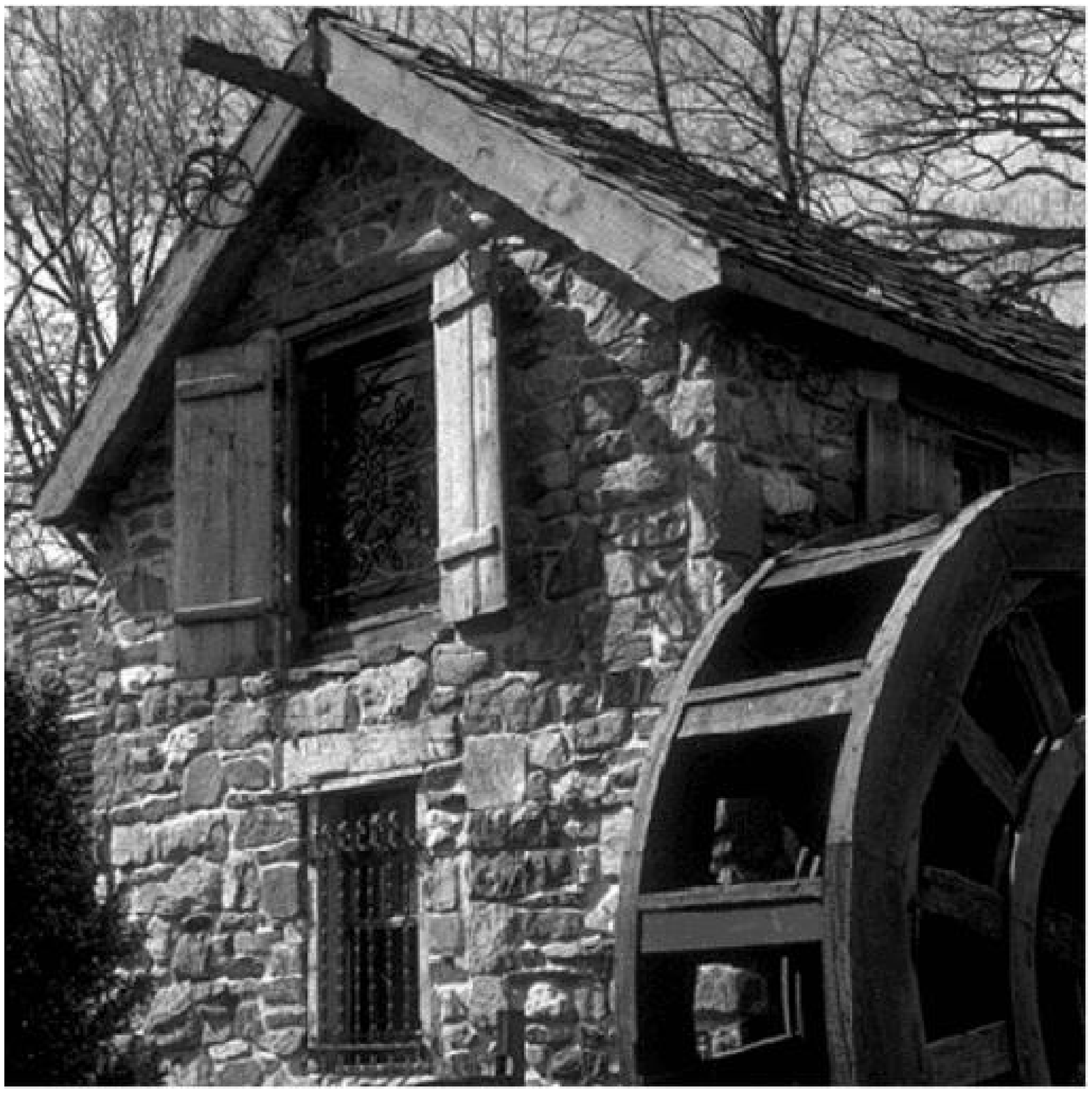}
	\label{exp:house:orig}
}
\subfigure[D4 orthogonal wavelet, compression 1:89, rectangular transform.] {
	\includegraphics[width=4cm]{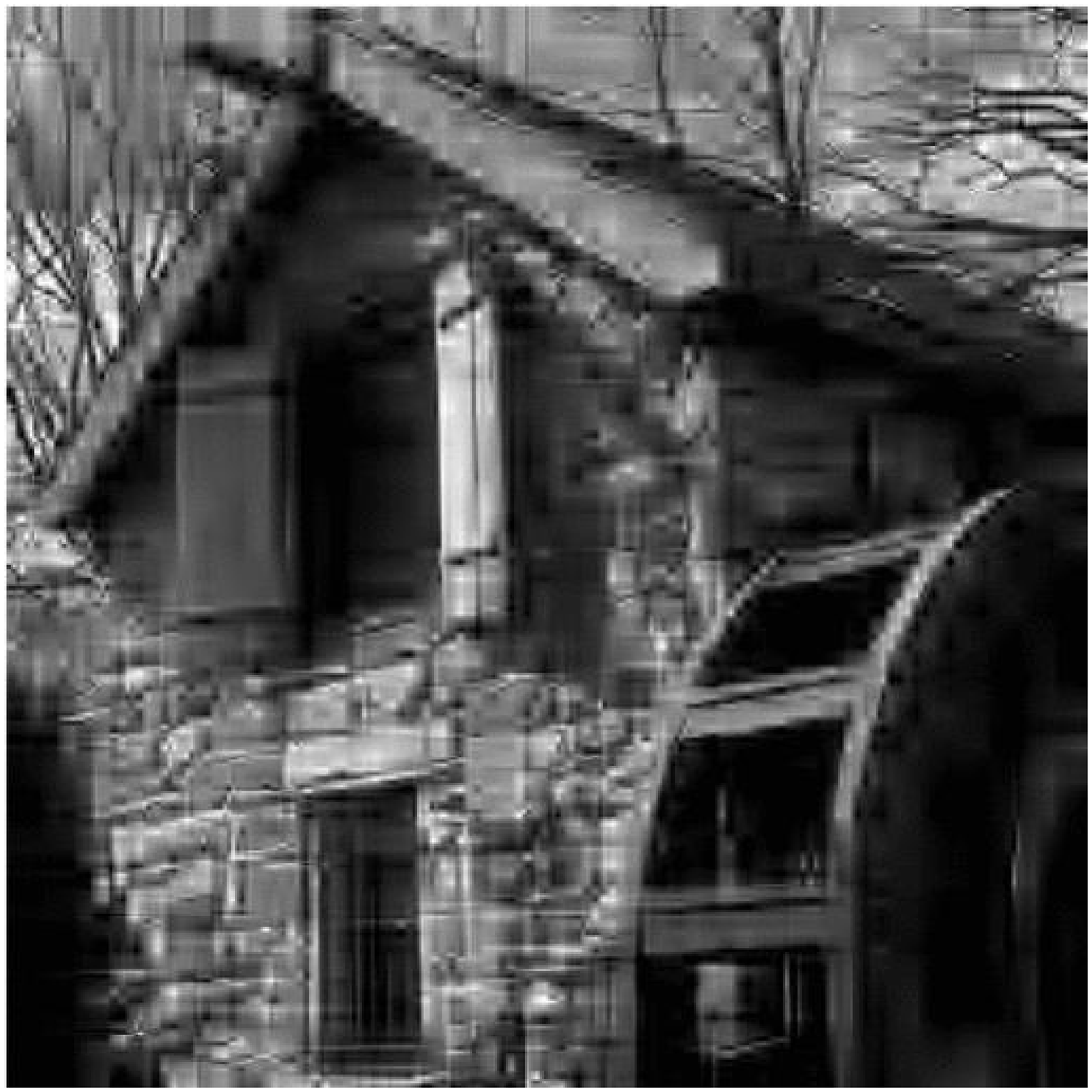}
	\label{exp:house:rect:d2:89}
}
\subfigure[D4 orthogonal wavelet, compression 1:86, square transform.] {
	\includegraphics[width=4cm]{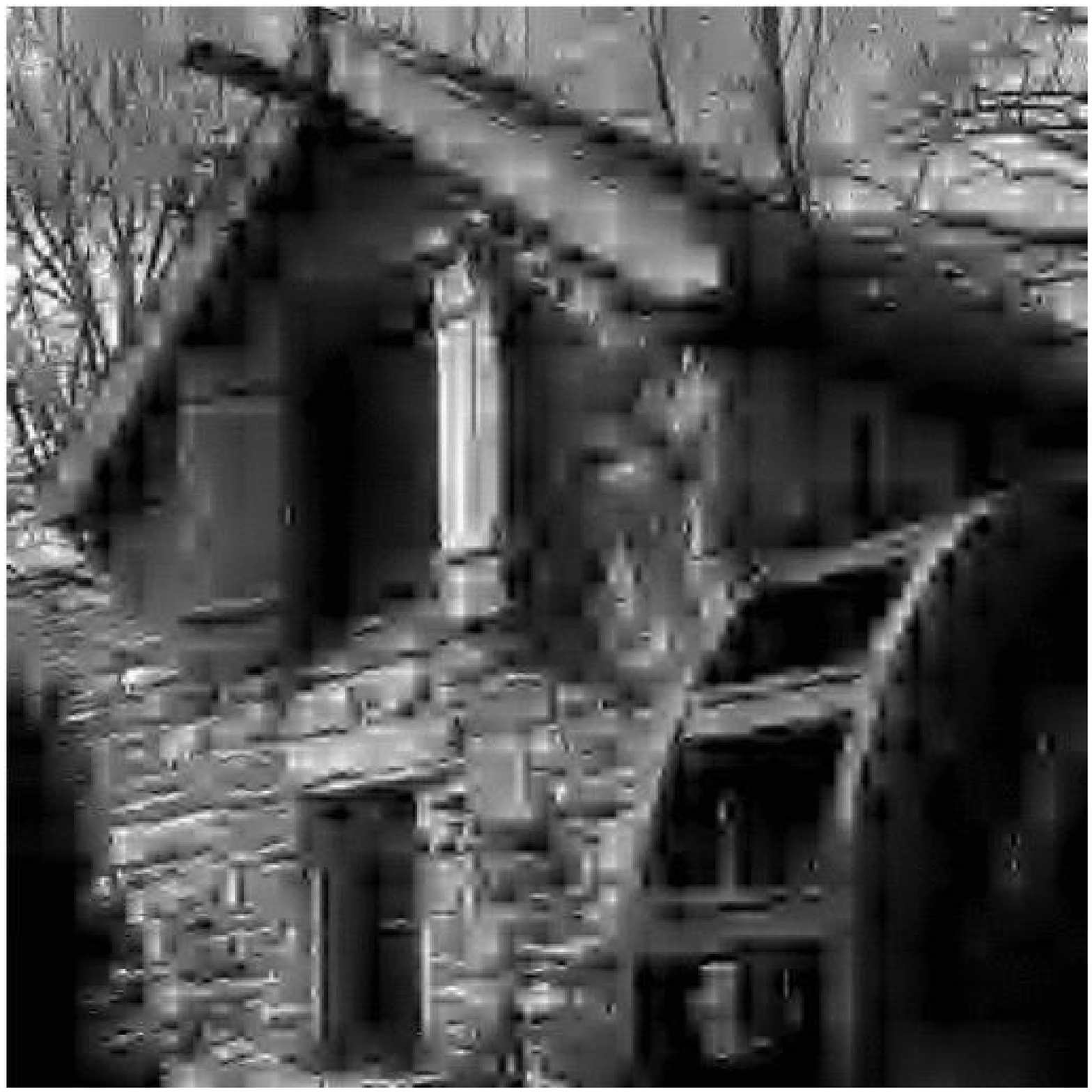}
	\label{exp:house:square:d2:86}
	}
\subfigure[CRF(13,7) biorthogonal wavelet, compression 1:82, rectangular transform.] {
	\includegraphics[width=4cm]{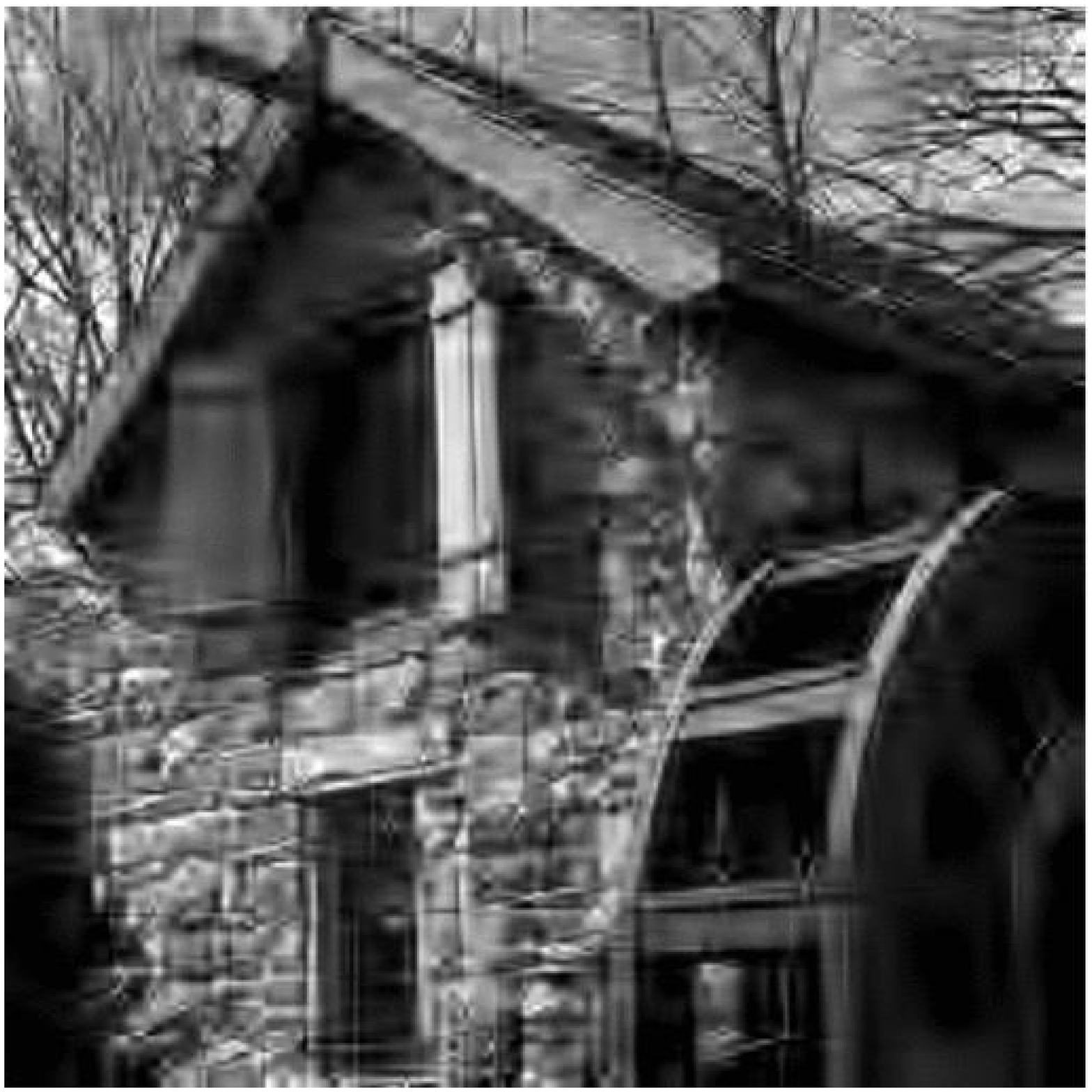}
	\label{exp:house:rect:crf137:79}
	}
\subfigure[CRF(13,7) biorthogonal wavelet, compression 1:79, square transform.] {
	\includegraphics[width=4cm]{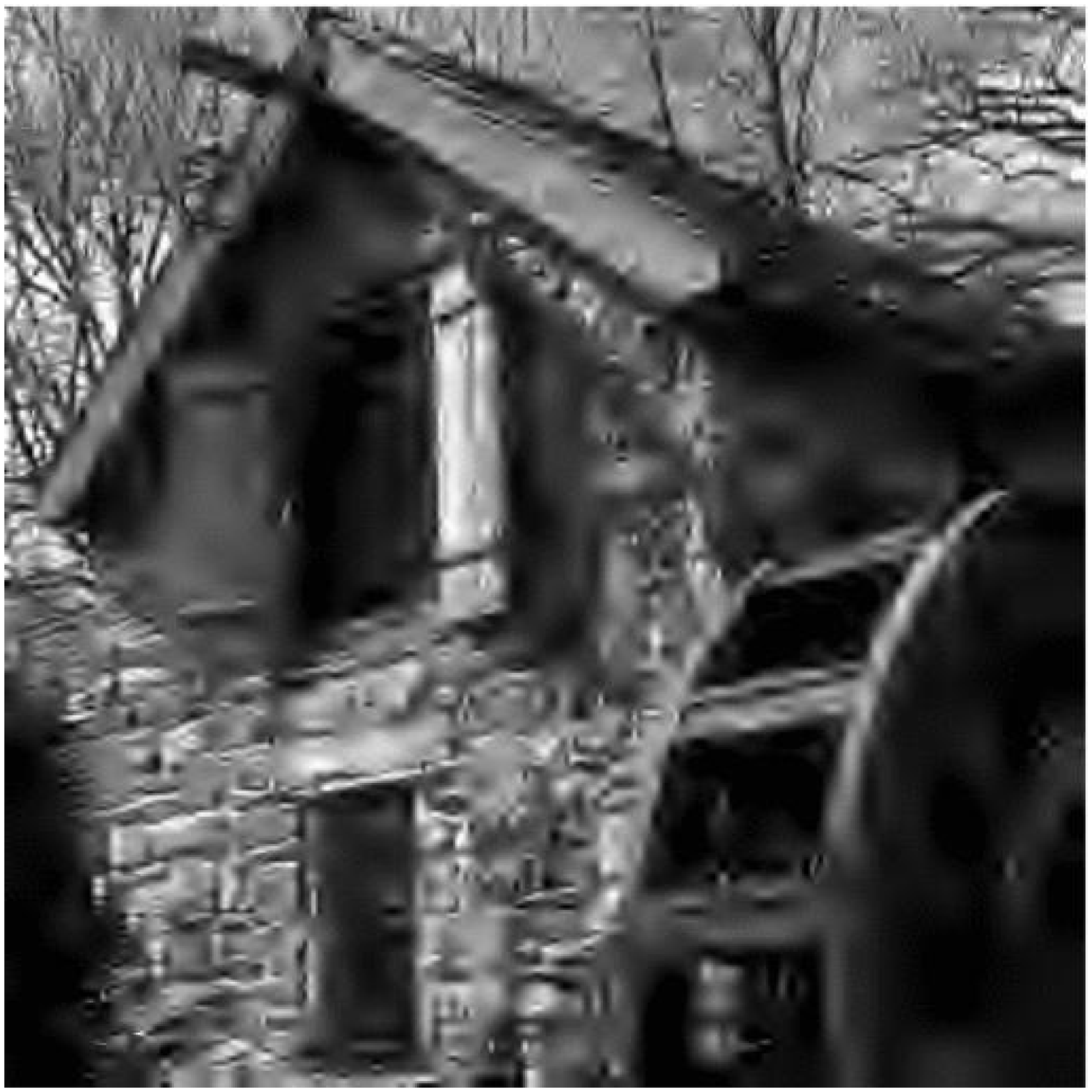}
	\label{exp:house:square:crf137:79}
	}
\subfigure[D4 orthogonal wavelet, compression 1:163, rectangular transform.] {
	\includegraphics[width=4cm]{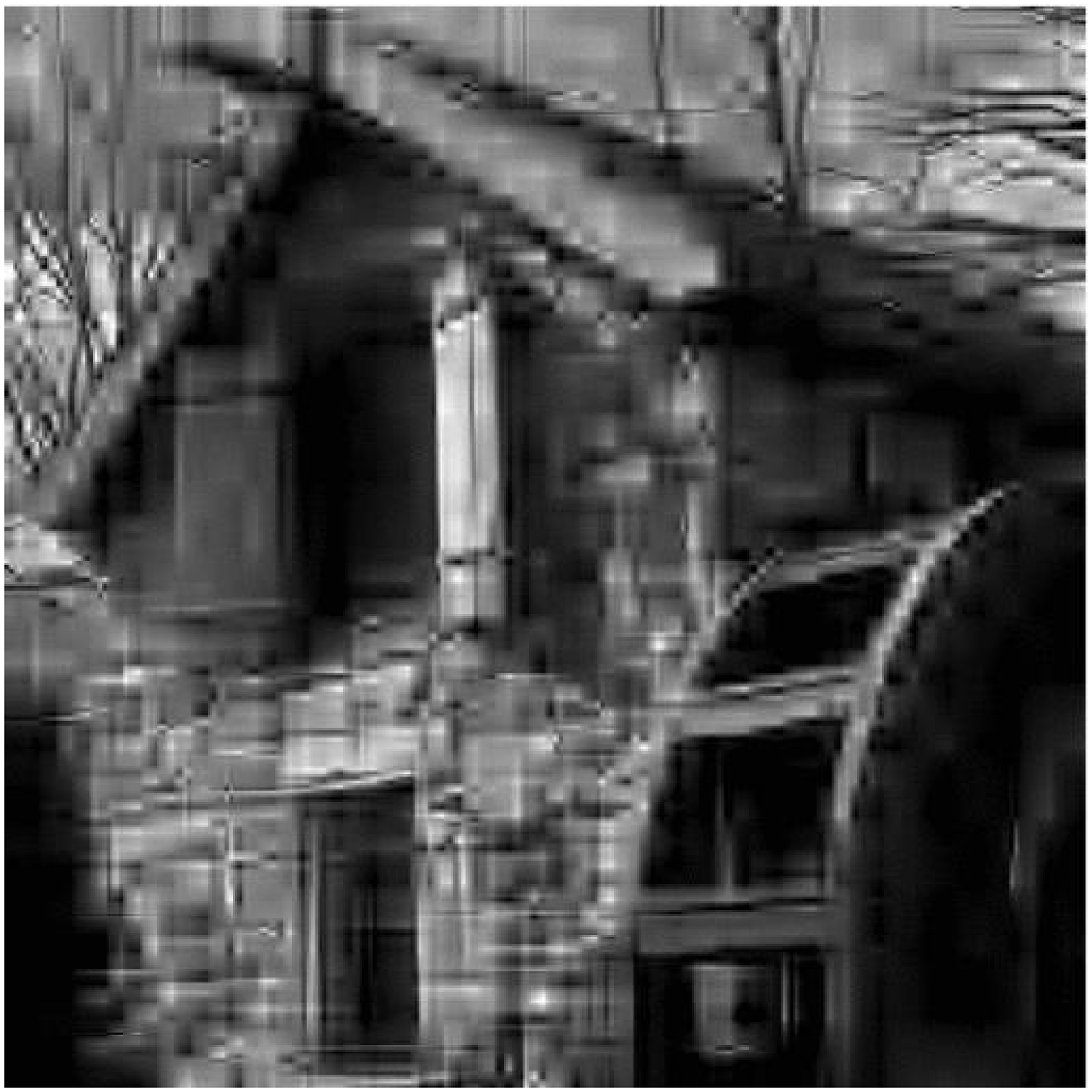}
	\label{exp:house:rect:d2:163}
	}
\subfigure[D4 orthogonal wavelet, compression 1:166, square transform.] {
	\includegraphics[width=4cm]{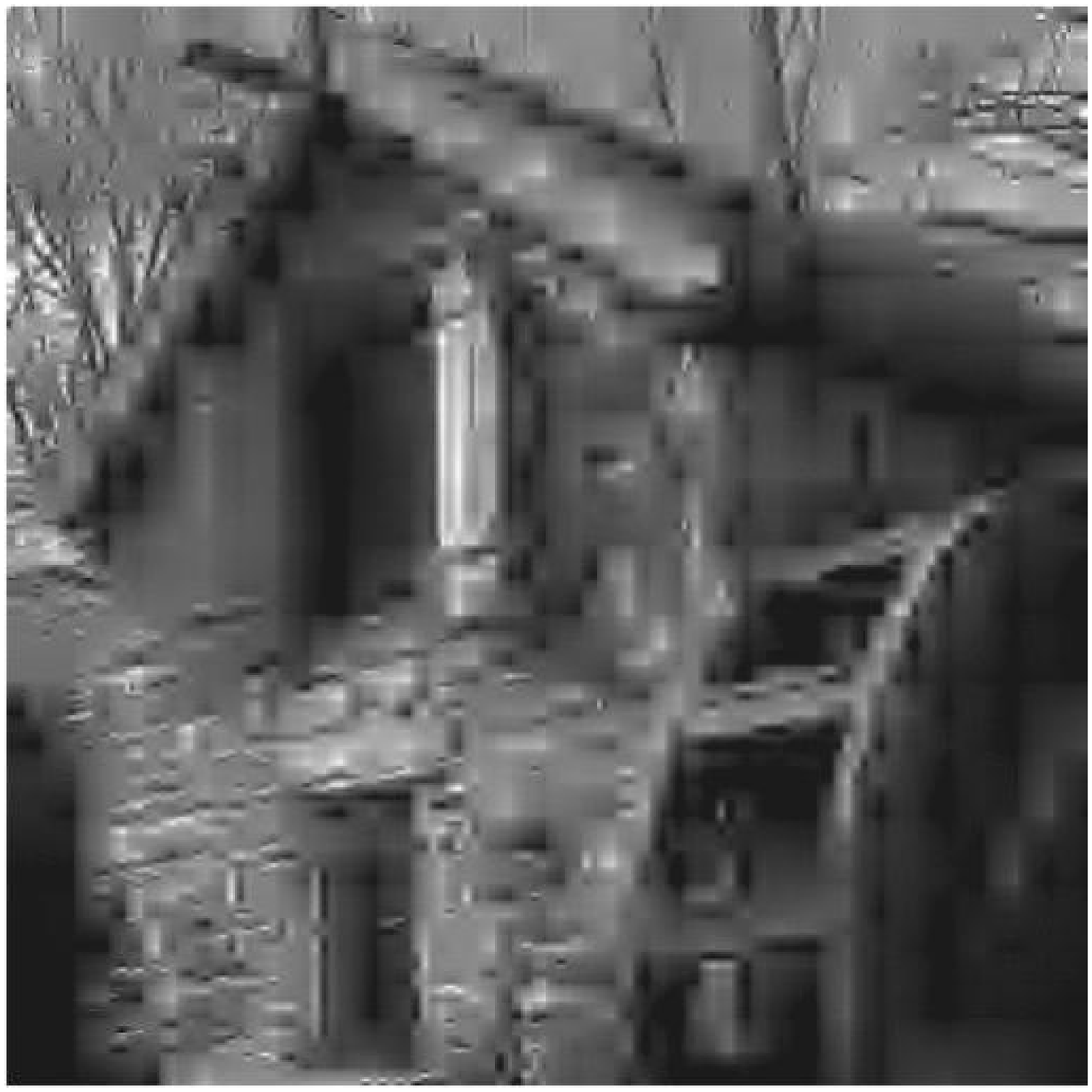}
	\label{exp:house:square:d2:166}
	}
\subfigure[CRF(13,7) biorthogonal wavelet, compression 1:163, rectangular transform.] {
	\includegraphics[width=4cm]{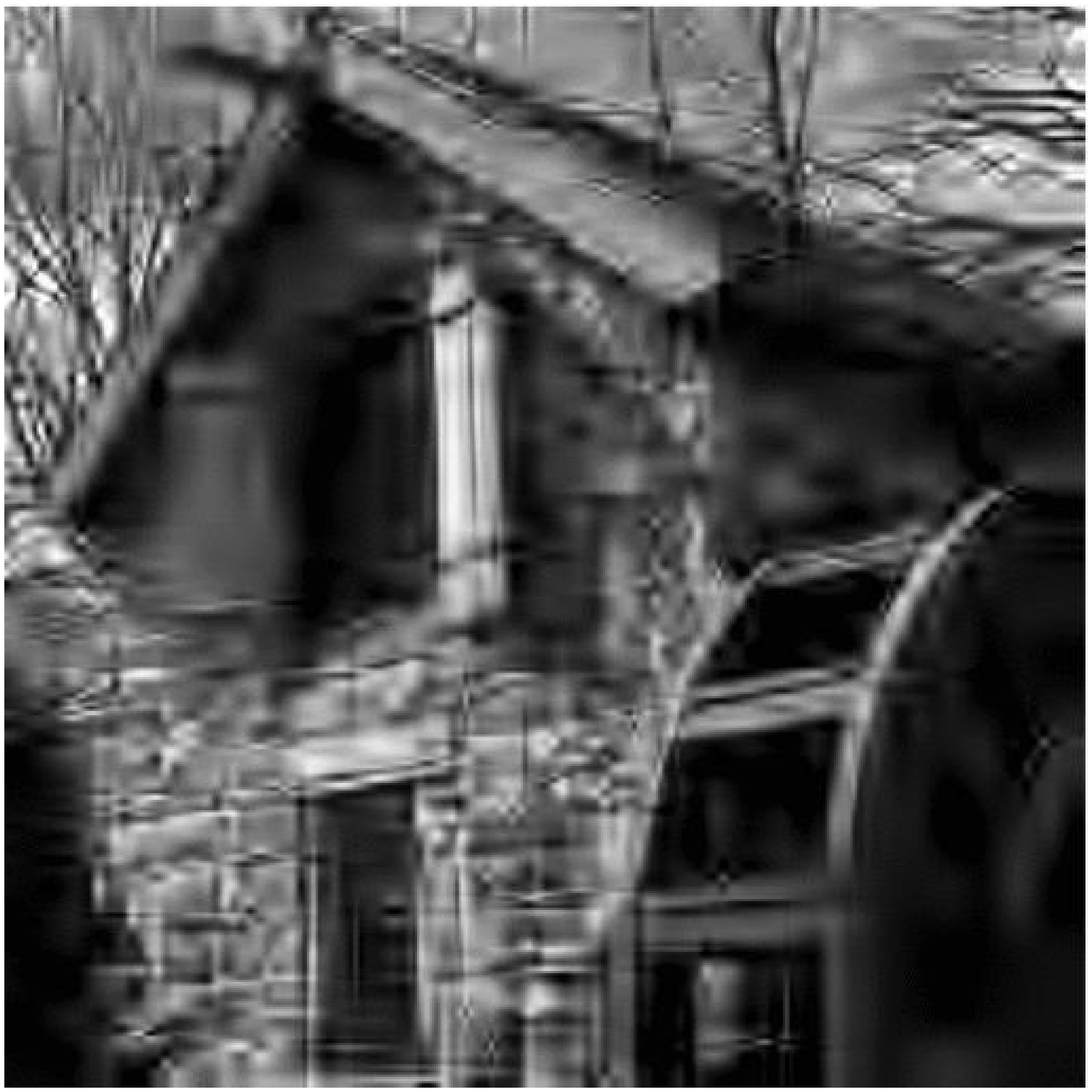}
	\label{exp:house:rect:crf137:163}
	}
\subfigure[CRF(13,7) biorthogonal wavelet, compression 1:162, square transform.] {
	\includegraphics[width=4cm]{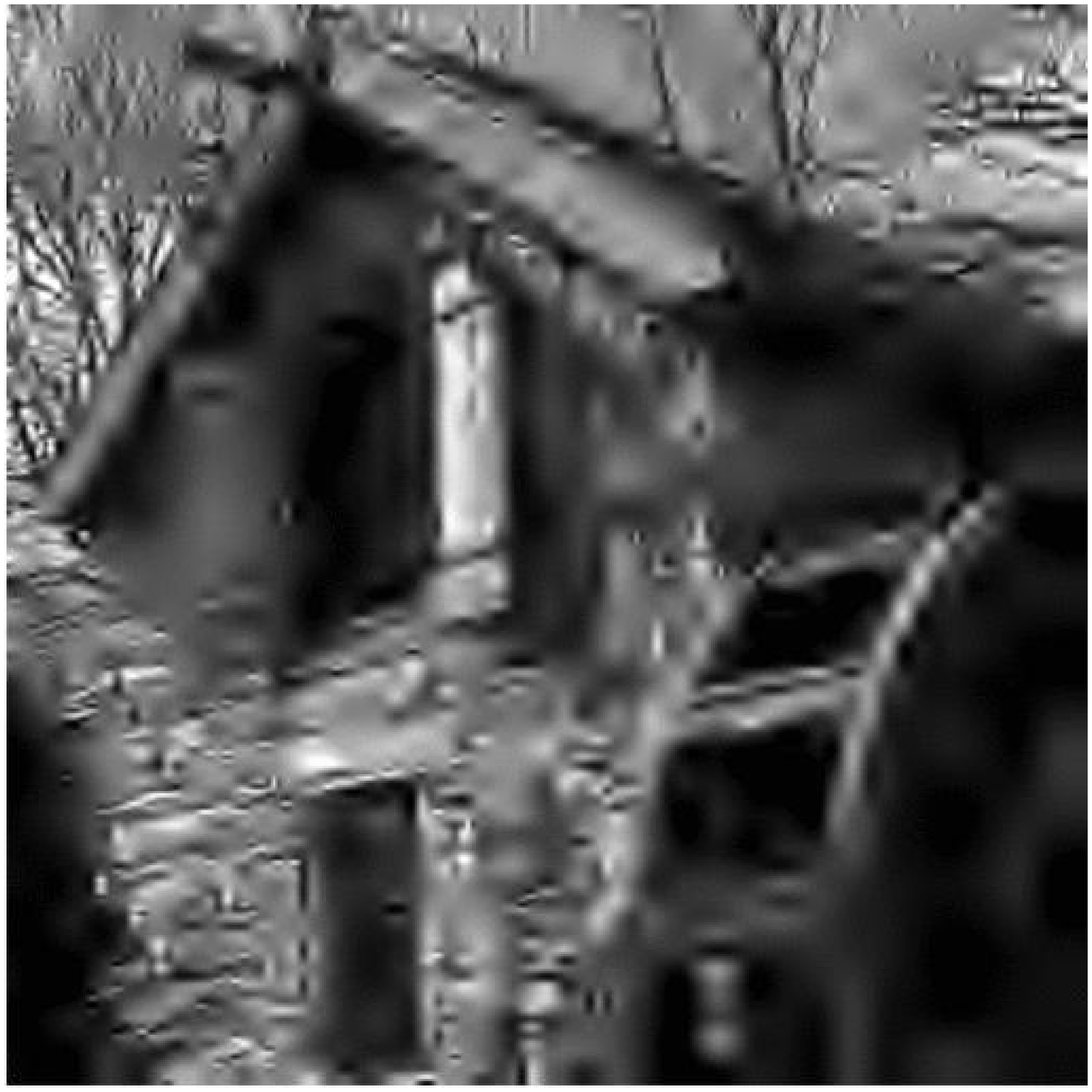}
	\label{exp:house:square:crf137:162}
	}
\caption{The "House"   image compressed by the rectangular and square wavelet transforms.}
\label{exp-house} 
\end{figure}

\begin{figure}[tbhp]
\centering
\subfigure[The original "Lena" $512\times 512$ image.] {
	\includegraphics[width=4cm]{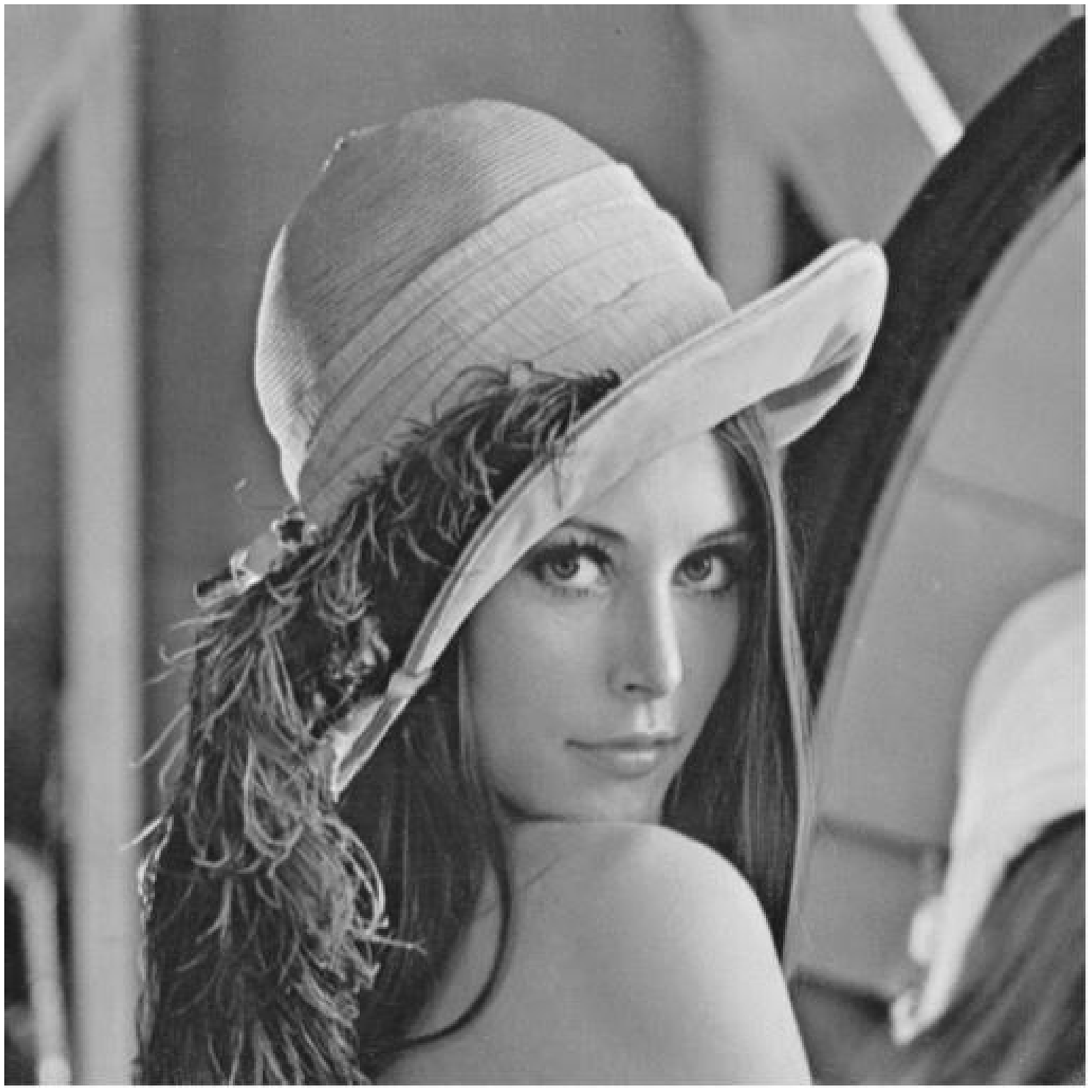}
	\label{exp:lena:orig}
}
\subfigure[D4 orthogonal wavelet, compression 1:87, rectangular transform.] {
	\includegraphics[width=4cm]{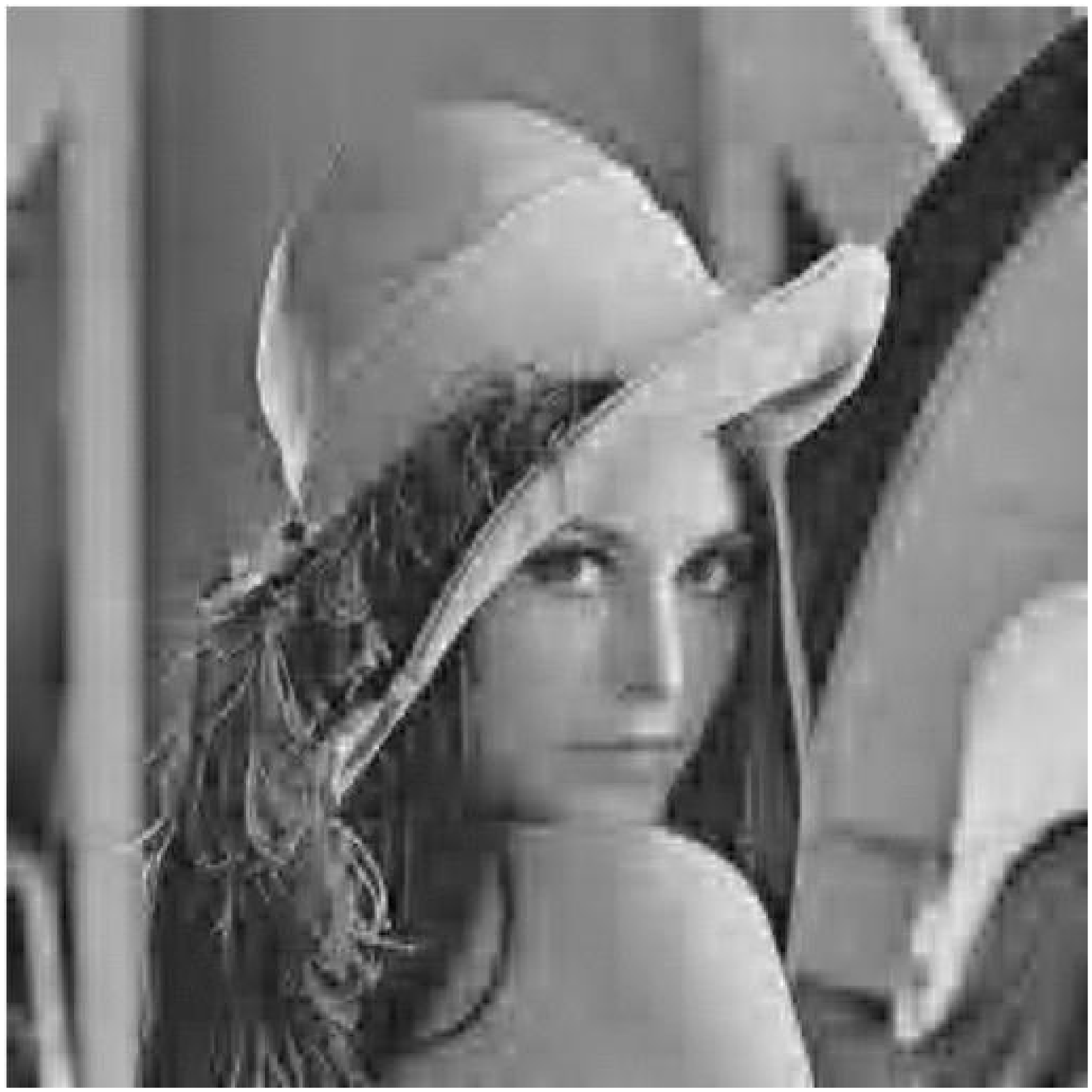}
	\label{exp:lena:rect:d2:87}
}
\subfigure[D4 orthogonal wavelet, compression 1:86, square transform.] {
	\includegraphics[width=4cm]{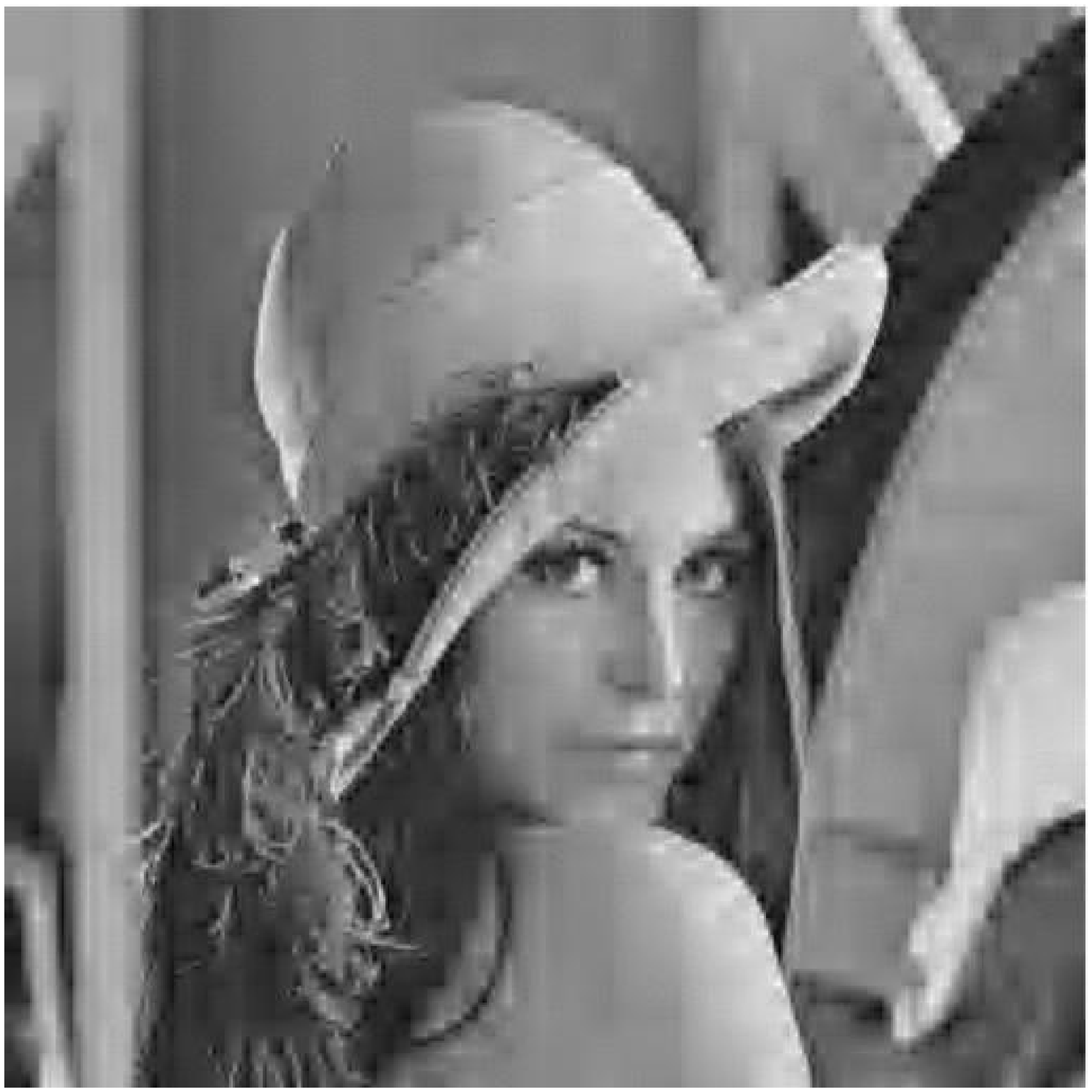}
	\label{exp:lena:square:d2:86}
	}
\subfigure[CRF(13,7) biorthogonal wavelet, compression 1:85, rectangular transform.] {
	\includegraphics[width=4cm]{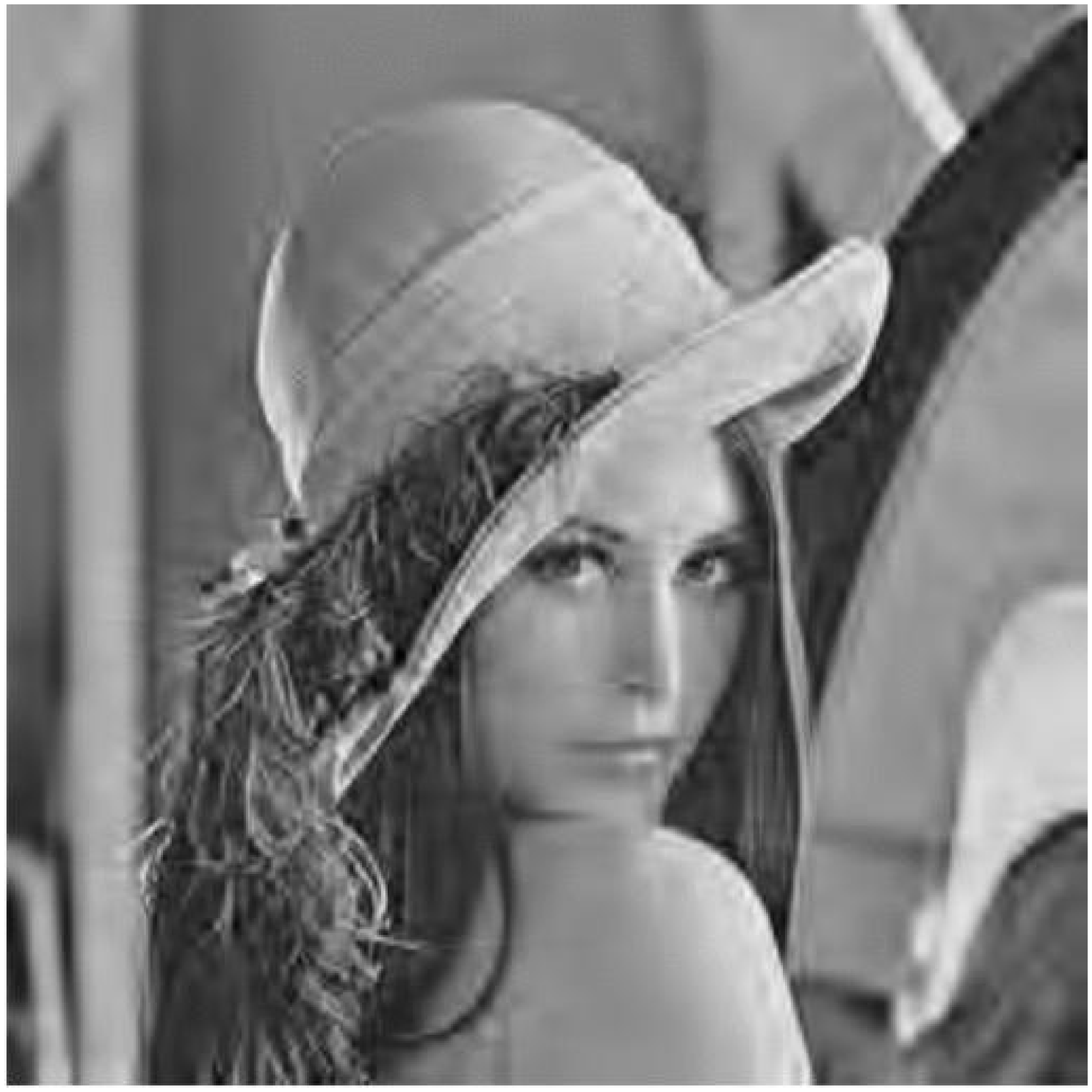}
	\label{exp:lena:rect:crf137:85}
	}
\subfigure[CRF(13,7) biorthogonal wavelet, compression 1:84, square transform.] {
	\includegraphics[width=4cm]{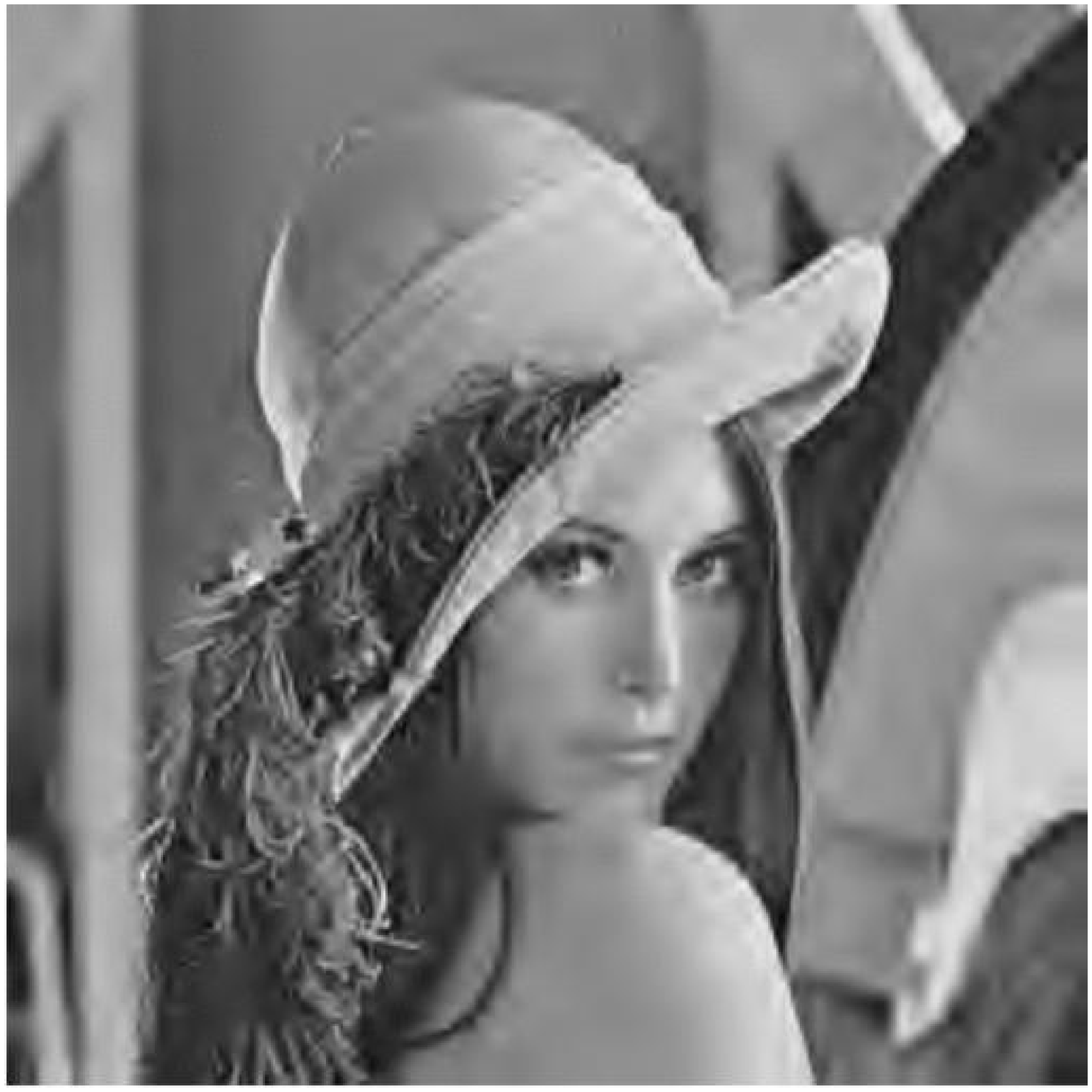}
	\label{exp:lena:square:crf137:84}
	}
\subfigure[D4 orthogonal wavelet, compression 1:160, rectangular transform.] {
	\includegraphics[width=4cm]{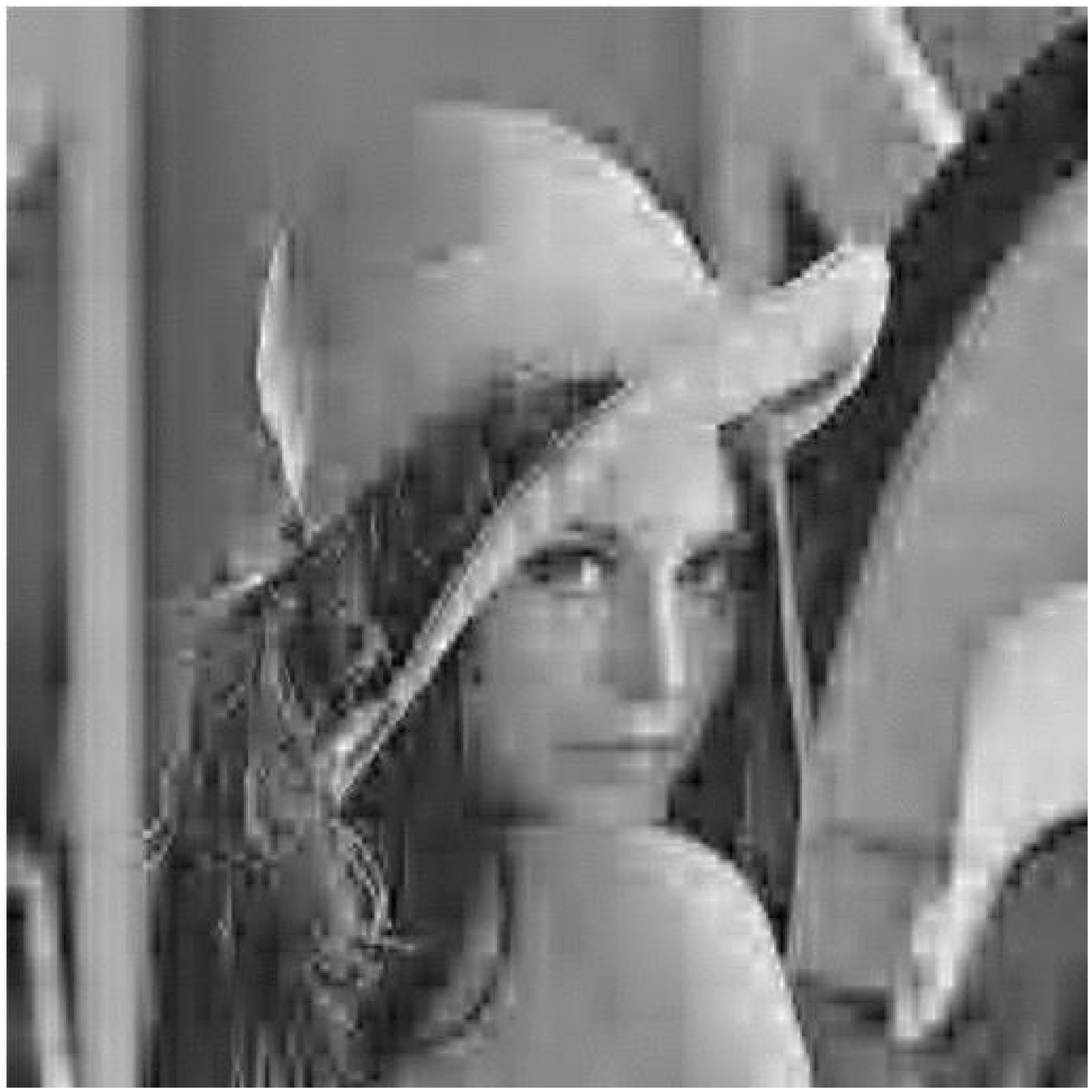}
	\label{exp:lena:rect:d2:160}
	}
\subfigure[D4 orthogonal wavelet, compression 1:166, square transform.] {
	\includegraphics[width=4cm]{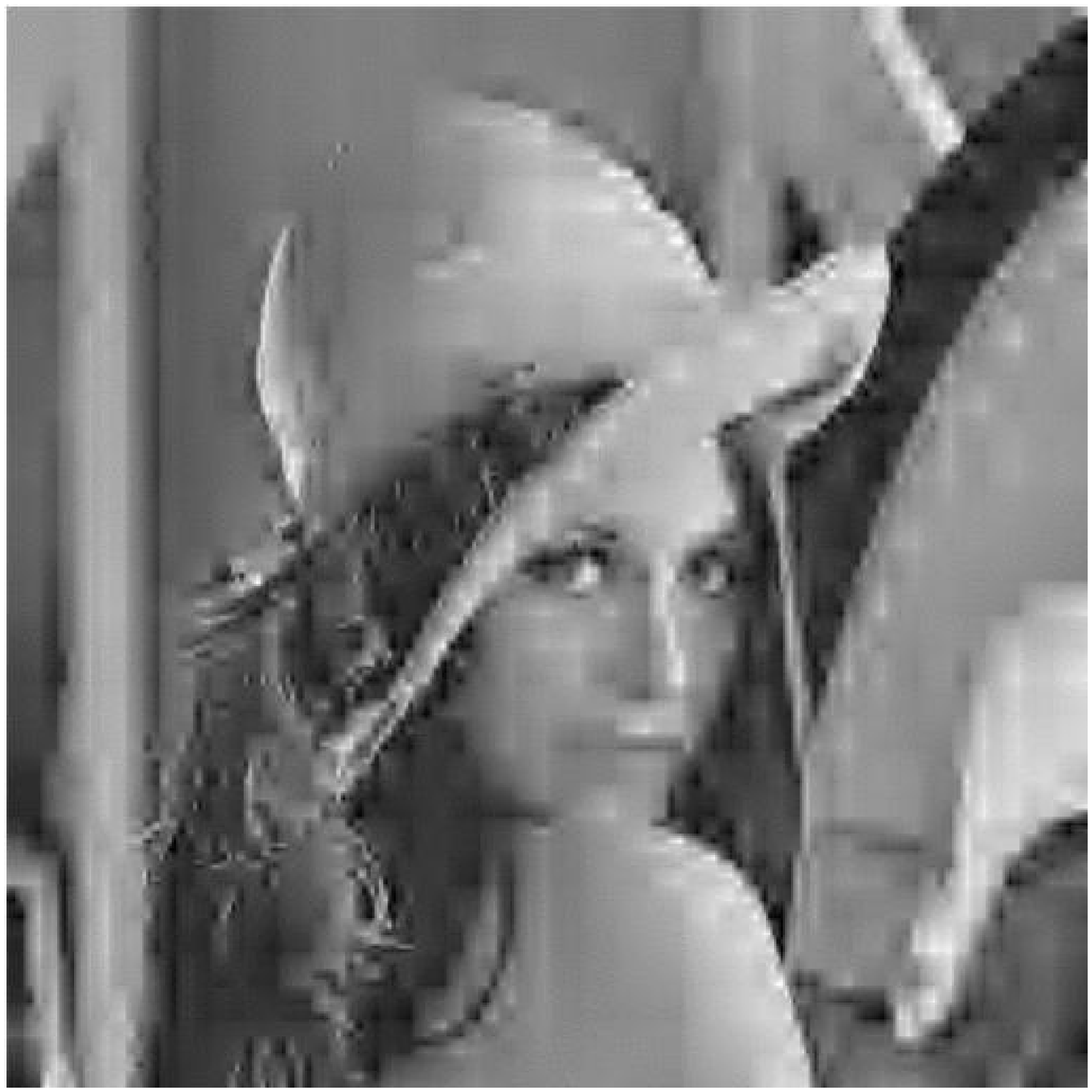}
	\label{exp:lena:square:d2:166}
	}
\subfigure[CRF(13,7) biorthogonal wavelet, compression 1:162, rectangular transform.] {
	\includegraphics[width=4cm]{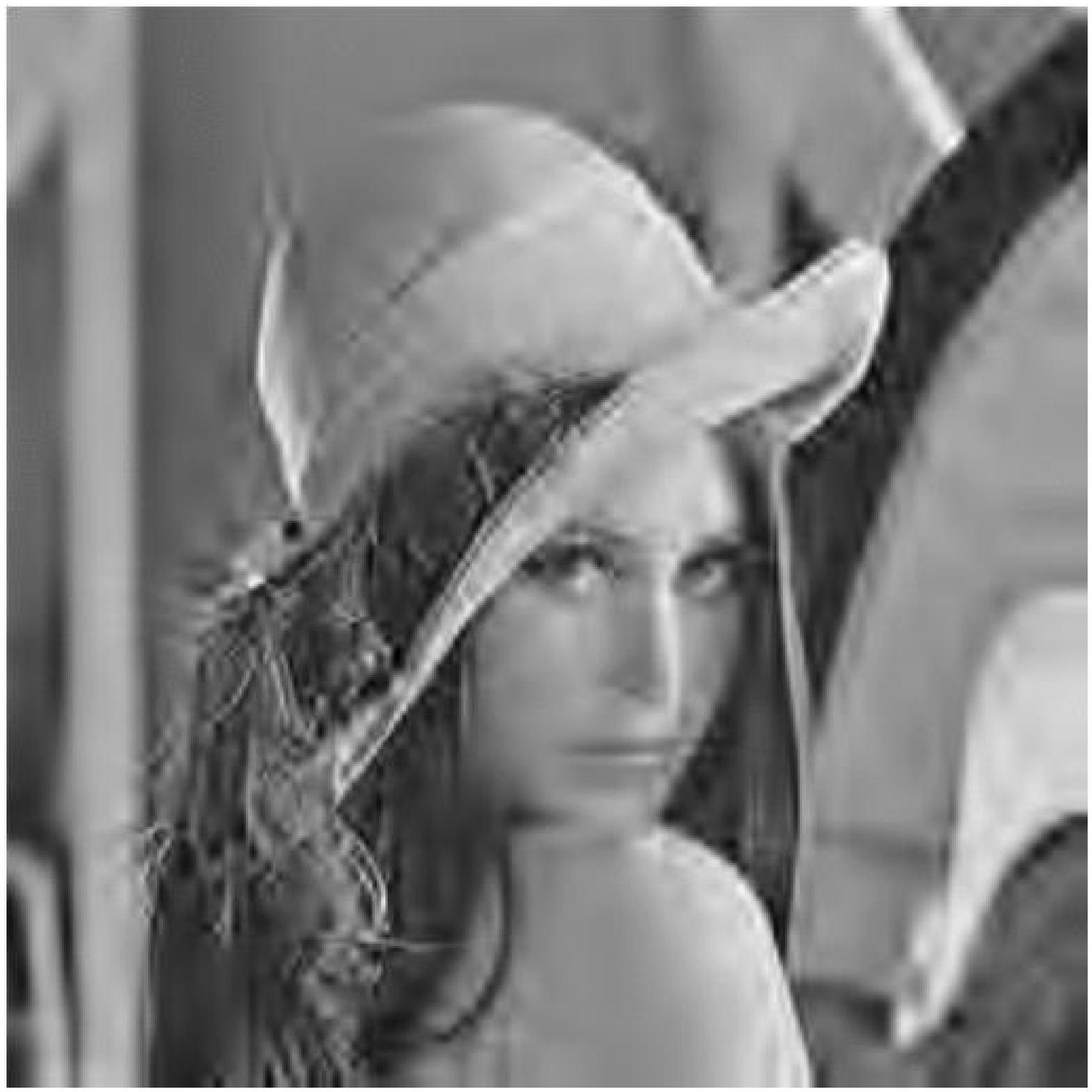}
	\label{exp:lena:rect:crf137:162}
	}
\subfigure[CRF(13,7) biorthogonal wavelet, compression 1:163, square transform.] {
	\includegraphics[width=4cm]{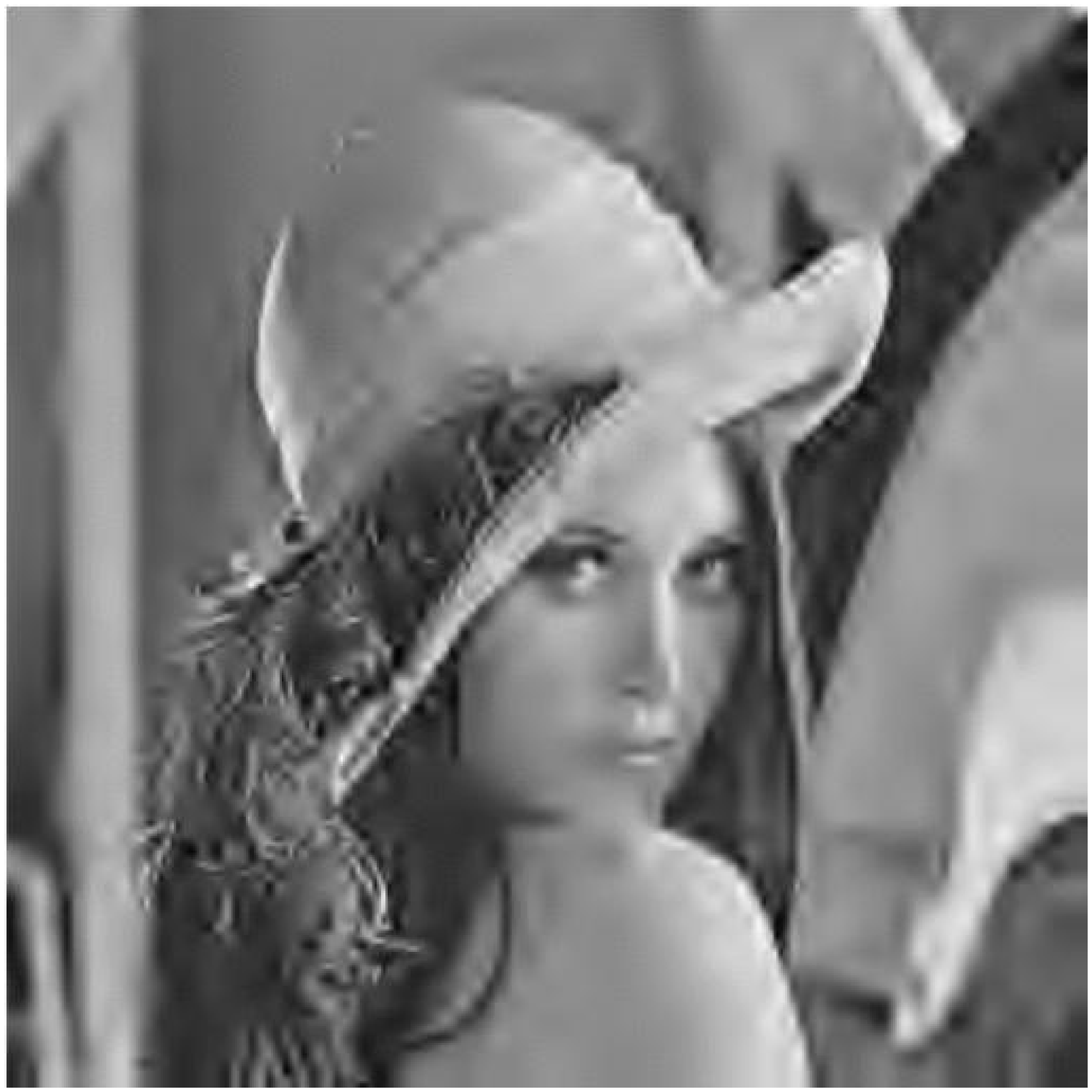}
	\label{exp:lena:square:crf137:163}
	}
\caption{The "Lena"   image compressed by the rectangular and square wavelet transforms.}
\label{exp-lena} 
\end{figure}
\begin{figure}[tbhp]
\centering
\subfigure[The original "Barbara" $512\times 512$ image.] {
	\includegraphics[width=4cm]{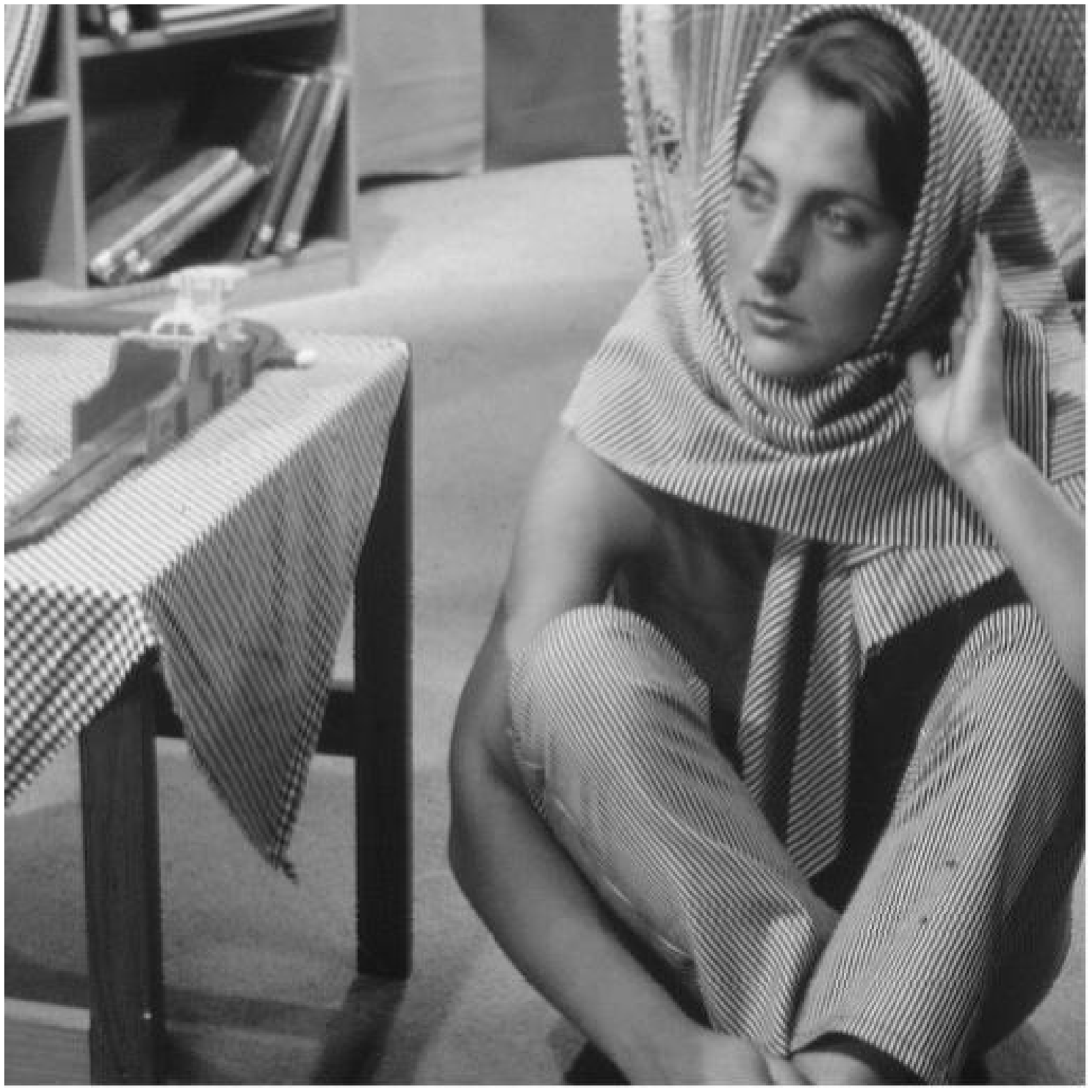}
	\label{exp:barbara:orig}
}
\subfigure[D4 orthogonal wavelet, compression 1:80, rectangular transform.] {
	\includegraphics[width=4cm]{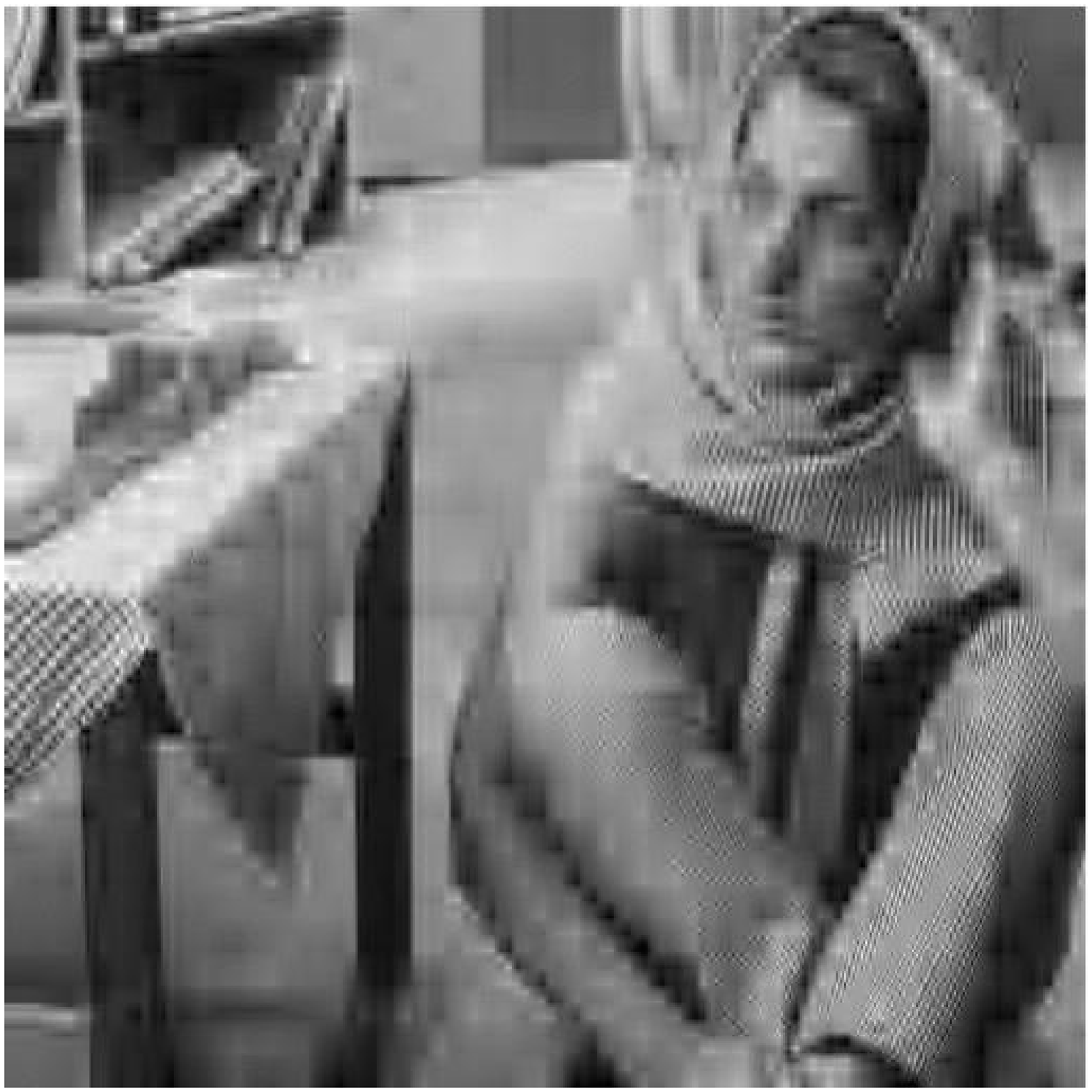}
	\label{exp:barbara:rect:d2:80}
}
\subfigure[D4 orthogonal wavelet, compression 1:92, square transform.] {
	\includegraphics[width=4cm]{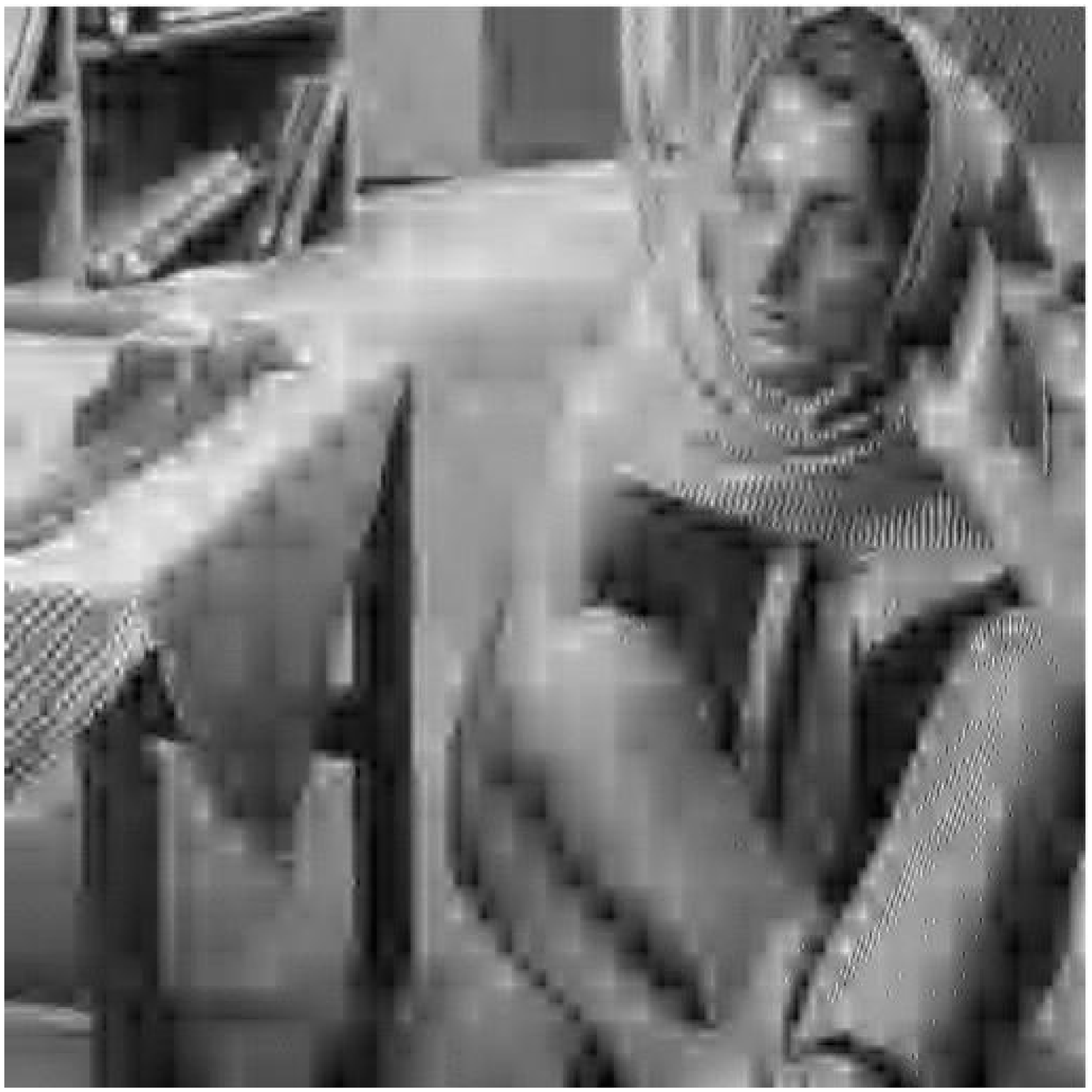}
	\label{exp:brabara:square:d2:92}
	}
\subfigure[CRF(13,7) biorthogonal wavelet, compression 1:88, rectangular transform.] {
	\includegraphics[width=4cm]{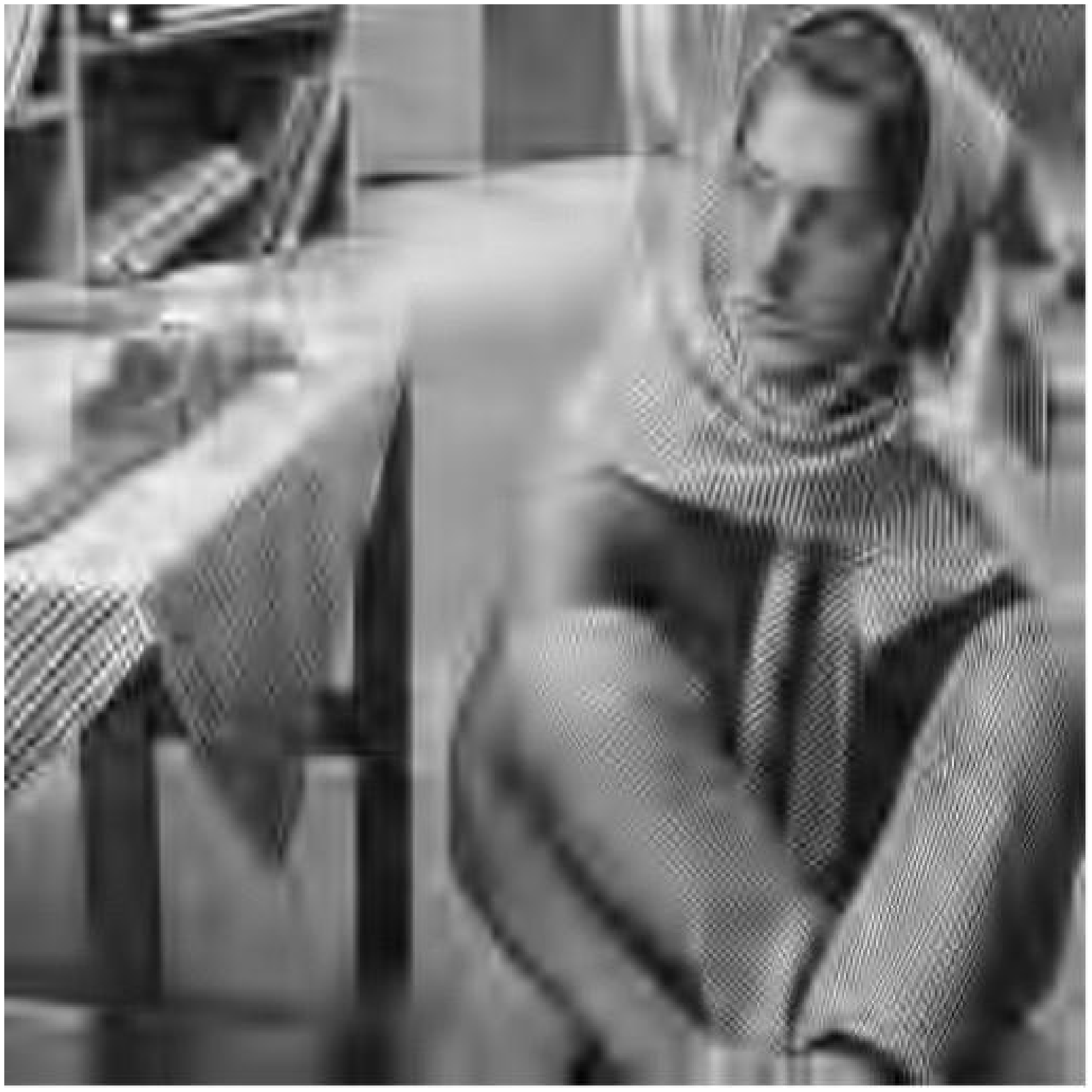}
	\label{exp:barbara:rect:crf137:88}
	}
\subfigure[CRF(13,7) biorthogonal wavelet, compression 1:83, square transform.] {
	\includegraphics[width=4cm]{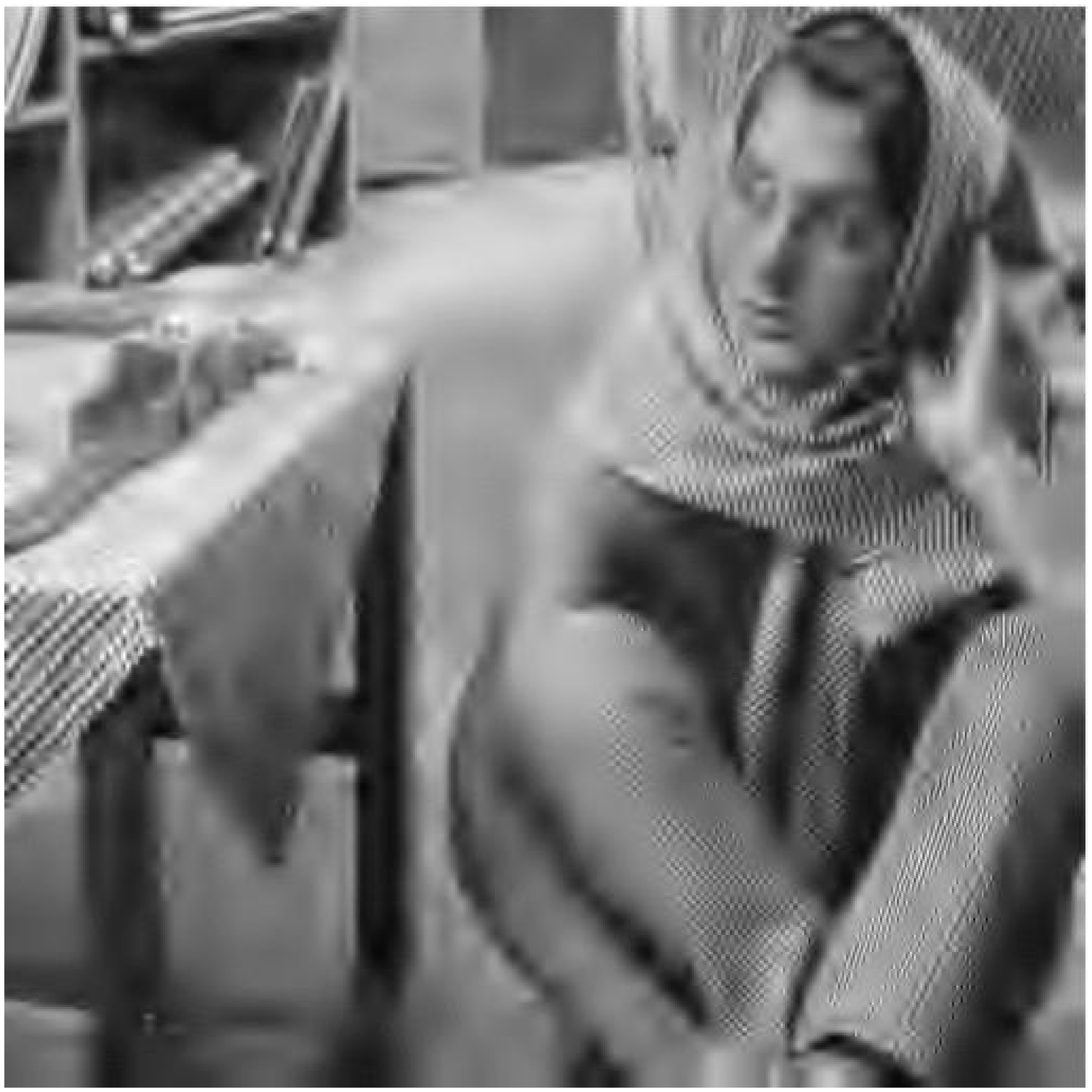}
	\label{exp:barbara:square:crf137:83}
	}
\subfigure[D4 orthogonal wavelet, compression 1:166, rectangular transform.] {
	\includegraphics[width=4cm]{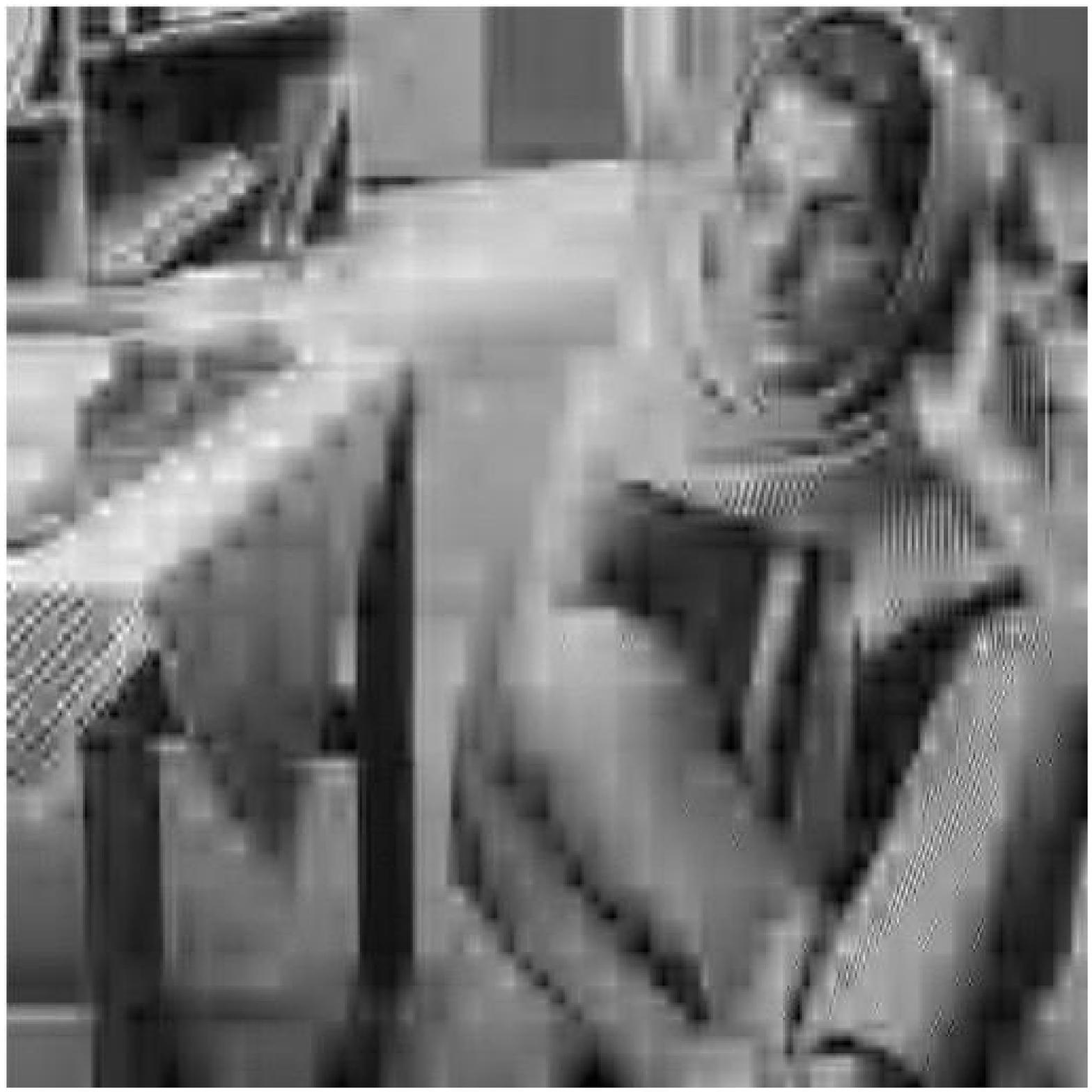}
	\label{exp:barbara:rect:d2:166}
	}
\subfigure[D4 orthogonal wavelet, compression 1:169, square transform.] {
	\includegraphics[width=4cm]{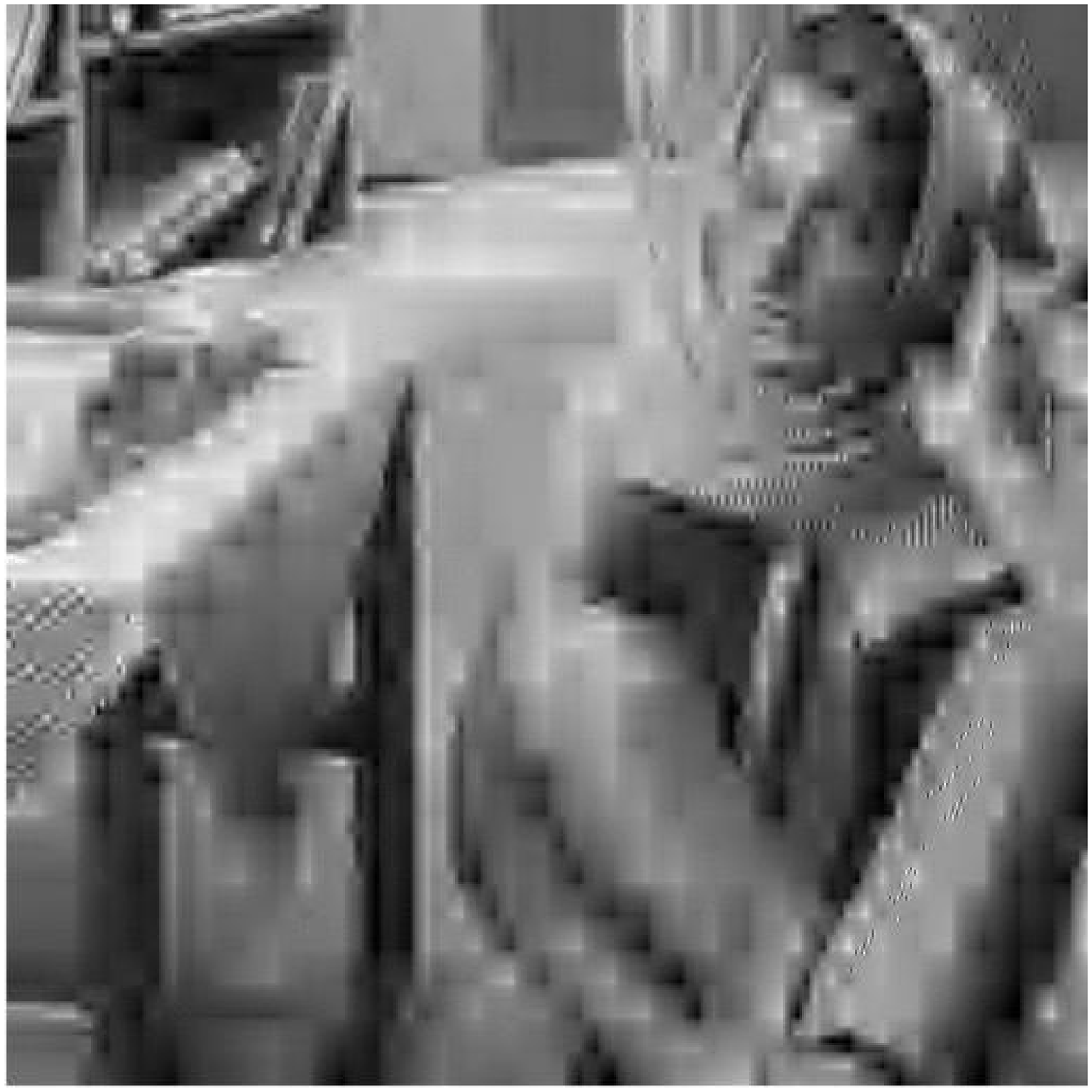}
	\label{exp:barbara:square:d2:169}
	}
\subfigure[CRF(13,7) biorthogonal wavelet, compression 1:166, rectangular transform.] {
	\includegraphics[width=4cm]{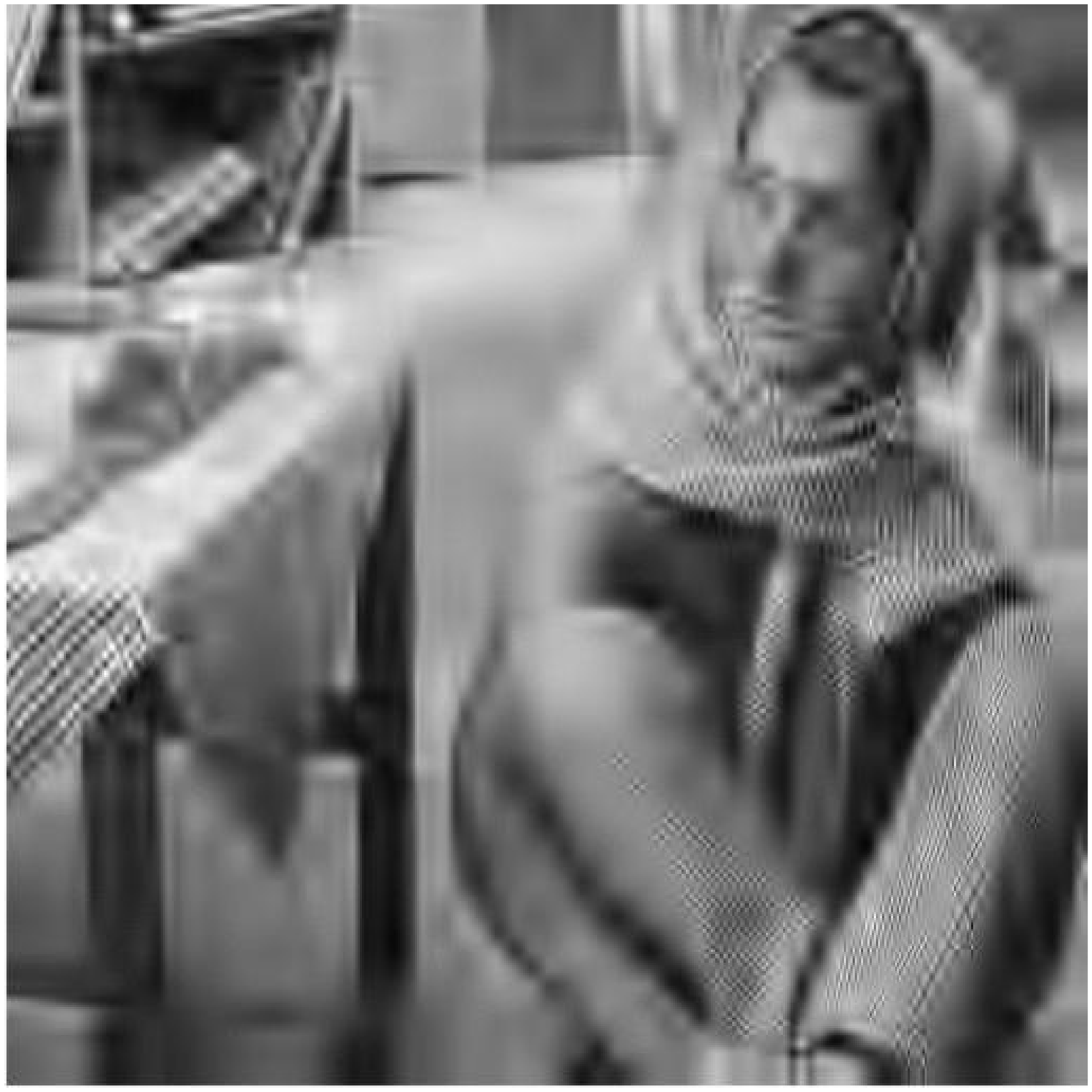}
	\label{exp:lena:rect:crf137:166}
	}
\subfigure[CRF(13,7) biorthogonal wavelet, compression 1:164, square transform.] {
	\includegraphics[width=4cm]{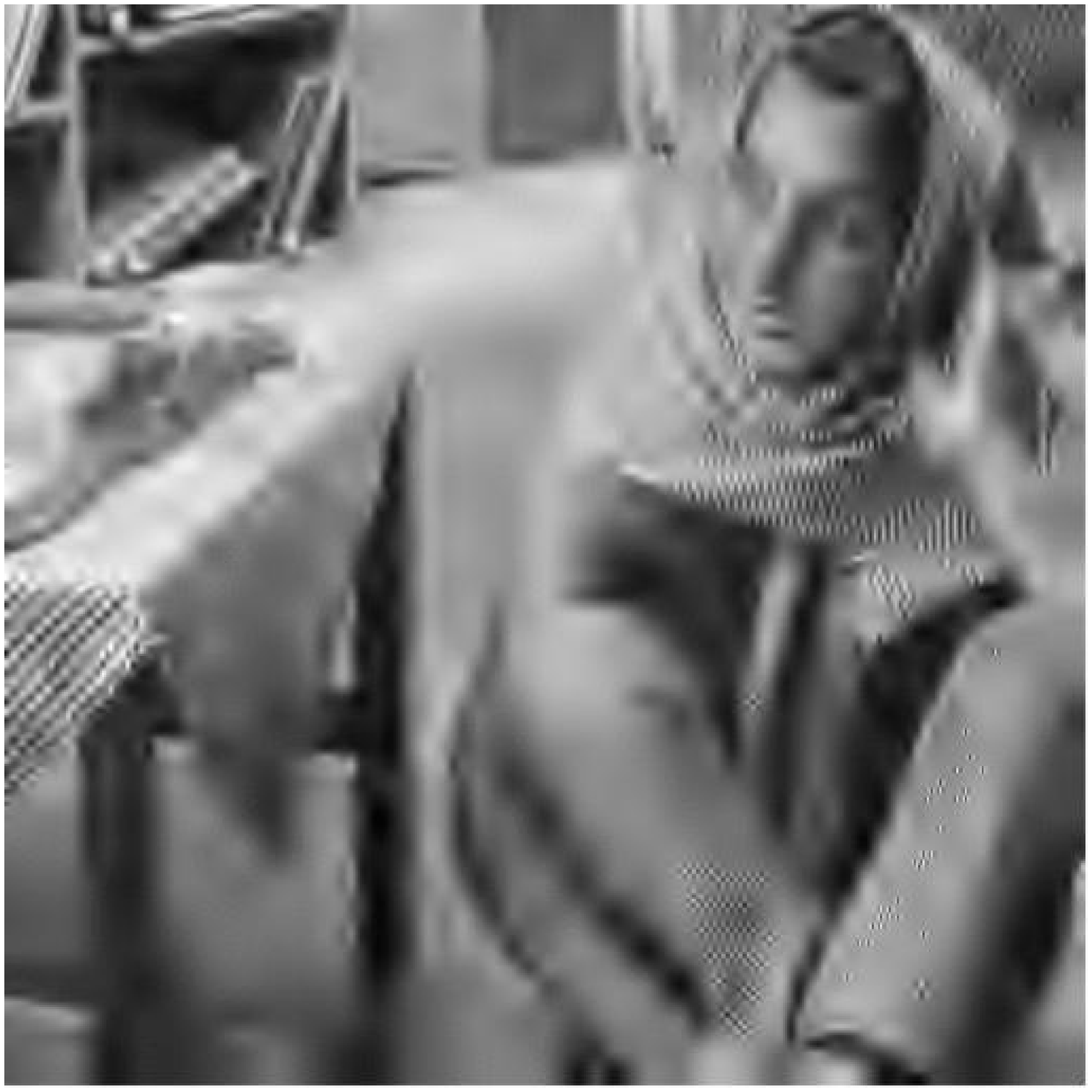}
	\label{exp:barbara:square:crf137:164}
	}
\caption{The "Barbara"   image compressed by the rectangular and square wavelet transforms.}
\label{exp-barbara} 
\end{figure}

To numerically compare results of square and rectangular wavelet transform we limited ourselves to the problem of image compression. We took the standard $512\times 512$ ``Mandrill'', ``House'', ``Lena'', and ``Barbara'' greyscale images and compress them to reduce  the number of non-zero coefficients by factors of $80$ and $160$\footnote{Some papers on image compression use $256 \times 256$ images. For the same quality, they report $4$ times smaller compression numbers.}.  The results are shown in Figs \ref{exp-mandrill}, \ref{exp-house}, \ref{exp-lena}, and \ref{exp-barbara}.

We believe that results  clearly show that the rectangular wavelet transform visually outperforms the square one for all test images. Although the effect is more visible on the  Mandrill and Barbara images and less visible on the  Lena  and House images, in all  cases, the results of rectangular compression at the 1:160 rate  are visually close to the results of square compression at the 1:80 rate.

\section{Conclusions}
\label{sec:conclusions}

As we shown, the rectangular wavelet transform has substantially better convergence rate if the approximated function has the mixed derivative of appropriate order. In this case, it allows to solve the "curse of dimensionality" issue and gives the same  degree of approximation (up to a logarithmic factor) as the wavelet basis in the one dimensional case.

From the results of numerical experiments we can see that the mixed derivative assumption is reasonable for the standard test images and rectangular wavelet transform allows to improve the compression rate (measured in the number of non-zero coefficient) by factor of 2.

It would be worthwhile to investigate what actual bit rate compression an  industrial grade image compression algorithm (such as JPEG2000) will produce if the square wavelet transform is replaced by rectangular one.

Although further experiments in this direction are needed, we believe that it's worthwhile to investigate the applicabity of rectangular wavelet transform to video compression. Because the actual dimension of data in this case is $3$,  the solution to the "curse of dimensionality" can be particularly beneficial in this case. Although certain technical work to organize codec to ensure intermediate frame recovery is needed, the results can be quite interesting not only from the actual bit-rate standpoint, but also because expensive movement search operation is replaced by a straightforward fast wavelet decomposition.

The rectangular wavelet transform is especially efficient to represent details that are either in the horizontal or vertical directions. Certain second generation schemes (such as wavelets for the quincunx lattice) can be modified in  to ensure the same rate of convergence for the functions with bounded mixed derivative as in the case of the rectangular wavelet transform. This may further improve the results.

\section{Appendix: proof of Lemma \ref{lemma:combined:bound}}
Because we assumed that $\tilde\psi$ is compactly supported, there exists $L>0$ such that $|x|\ge L \Rightarrow \tilde\psi(x)=0$.
Let us define functions $\tilde\Psi_m$ for $m\in\mathbb{N}$ as follows:
\begin{equation}
\tilde\Psi_m(x) :=\int_{-L}^x\tau^{m-1}\tilde\psi(\tau)\,d\tau.
\end{equation}
Because $\tilde\psi\in L_\infty$ , $\tilde\Psi_m(x)\in L_\infty^{loc}$. 
Moreover, because $\tilde\psi$ has    $M$ vanishing moments $\tilde\Psi_m$  is supported on $[-L,L]$ for $m\le M$. 

Lemma \ref{lemma:combined:bound} is a combination of the following results:

\begin{lem}
For any function $f(x)$ with $M$-th derivative 
\begin{equation}
<f(x),\tilde\psi(\sigma x-\theta)> = \frac{1}{\sigma^M} \int_{\frac{\theta-L}{\sigma}}^{\frac{\theta+L}{\sigma}} f^{(M)}(x)\tilde\Psi_M(\sigma x-\theta)\, dx.
\label{coef:partial}
\end{equation}
\end{lem}
\begin{pf}
\begin{multline}
<f(x),\tilde\psi(\sigma x-\theta)> = \int_{\frac{\theta-L}{\sigma}}^{\frac{\theta+L}{\sigma}} f(x)\tilde\psi(\sigma x-\theta)\, dx =\\
=
\int_{\frac{\theta-L}{\sigma}}^{\frac{\theta+L}{\sigma}} f^{'}(x)\int_{\frac{\theta-L}{\sigma}}^x\tilde\psi(\sigma \tau-\theta)d\tau\, dx =
\cdots = \\ =
\int_{\frac{\theta-L}{\sigma}}^{\frac{\theta+L}{\sigma}} f^{(M)}(x)\int_{\frac{\theta-L}{\sigma}}^x \big[\tau-\frac{\theta-L}{\sigma}\big]^{M-1}\tilde\psi(\sigma \tau-\theta)d\tau\, dx. 
\end{multline}
\end{pf}
\begin{lem}
\label{lemma:bound} For any wavelet basis with $M$ dual vanishing moments and for any $p\ge 1$ there exist  constants $C,L$ such 
that for any $\sigma>0$ and $\theta\in\mathbb{R}$ and for any function $f$ with $M$-th derivative in $L_p^{loc}$ 
\begin{equation}
\big | <f(x),\tilde\psi(\sigma x-\theta)>\big|\le 
C \frac{1}{\sigma^{M+1-1/p}}\left[\int_{\frac{\theta-L}{\sigma}}^{\frac{\theta+L}{\sigma}} |f^{(M)}(x)|^p\, dx\right]^{1/p}. 
\end{equation}
Moreover, if the basis does not has $M+1$ dual vanishing moments, then there exists a function $f$ such that
\begin{equation}
\big | <f(x),\tilde\psi(\sigma x-\theta)>\big|\ge  
\frac{1}{C} \frac{1}{\sigma^{M+1-1/p}}\left[\int_{\frac{\theta-L}{\sigma}}^{\frac{\theta+L}{\sigma}} |f^{(M)}(x)|^p\, dx\right]^{1/p}. 
\label{bound:coef:inverse}
\end{equation}
\end{lem}
\begin{pf}
Denote $\mathcal{N}_q:=\| \tilde\Psi_M\|_{L_q}$. Let us note that $\mathcal{N}_q<\infty$ for $q>0$.

By applying the H\"older inequality to \eqref{coef:partial}, we get
\begin{equation}
\big | <f(x),\tilde\psi(\sigma x-\theta)>\big|\le  \frac{\mathcal{N}_{1/(1-1/p}}{\sigma^{M+1-1/p}}\left[\int_{\frac{\theta-L}{\sigma}}^{\frac{\theta+L}{\sigma}} |f^{(M)}(x)|^p\, dx\right]^{1/p}. 
\end{equation}  
To proof \eqref{bound:coef:inverse},  consider  $f(x)=x^M/M!$. After  substituting it to \eqref{coef:partial}:
\begin{equation}
<\frac{x^M}{M!},\tilde\psi(\sigma x-\theta)> = \frac{1}{\sigma^{M+1}} \tilde\Psi_{M+1}(L).
\end{equation}
Note that since $\tilde\psi$ does not has $M+1$ vanishing moments, $\tilde\Psi_{M+1}(L)\ne 0$.
 Taking into the account that 
\begin{equation}
\left[\int_{\frac{\theta-L}{\sigma}}^{\frac{\theta+L}{\sigma}} 1 \, dx\right]^{1/p}= (2L/\sigma)^{1/p},
\end{equation}
and putting $C=\max((2L)^{1/p}/|\tilde\Psi_{M+1}(L)|, \mathcal{N}_{1/(1-1/p})$, we conclude the proof.
\end{pf}
\begin{cor}
For any wavelet basis with $M$ dual vanishing moments and for any $p\ge 1$ there exist  constants $C,L$ such 
that for any $\sigma_1,\sigma_2>0$ and $\theta_1,\theta_2\in\mathbb{R}$ and for any function $f(x,y)$ with the appropriate  derivative in $L_p^{loc}$:
\begin{multline}
\big | <f(x,y),\tilde\psi(\sigma_1 x-\theta_1)\tilde\phi(\sigma_2 y-\theta_2)>\big|\le \\
\le C \frac{1}{\sigma_1^{M+1-1/p}}\left[
\int_{\frac{\theta_1-L}{\sigma_1}}^{\frac{\theta_1+L}{\sigma_1}}
\int_{\frac{\theta_2-L}{\sigma_2}}^{\frac{\theta_2+L}{\sigma_2}}
 |\frac{d^M}{dx^M}f(x,y)|^p\, dy\, dx\right]^{1/p},
\end{multline}
\begin{multline}
\big | <f(x,y),\tilde\phi(\sigma_1 x-\theta_1)\tilde\psi(\sigma_2 y-\theta_2)>\big|\le \\
\le C \frac{1}{\sigma_2^{M+1-1/p}}\left[
\int_{\frac{\theta_1-L}{\sigma_1}}^{\frac{\theta_1+L}{\sigma_1}}
\int_{\frac{\theta_2-L}{\sigma_2}}^{\frac{\theta_2+L}{\sigma_2}}
 |\frac{d^M}{dy^M}f(x,y)|^p\, dy\, dx\right]^{1/p}. 
 \end{multline}
\end{cor} 
\begin{lem}
For any wavelet basis with $M$ dual vanishing moments and for any $p\ge 1$ there exist  constants $C,L$ such 
that for any $\sigma_1,\sigma_2>0$ and $\theta_1,\theta_2\in\mathbb{R}$ and for any function $\frac{d^{2M}}{dx^Mdy^M}f(x,y)\in L_p^{loc}$:
\begin{multline}
\big | <f(x,y),\tilde\psi(\sigma_1 x-\theta_1)\tilde\psi(\sigma_2 y-\theta_2)>\big|\le \\
\le C \frac{1}{(\sigma_1\sigma_2)^{M+1-1/p}}\left[
\int_{\frac{\theta_1-L}{\sigma_1}}^{\frac{\theta_1+L}{\sigma_1}}
\int_{\frac{\theta_2-L}{\sigma_2}}^{\frac{\theta_2+L}{\sigma_2}}
 |\frac{d^{2M}}{dx^Mdy^M}f(x,y)|^p\, dy\, dx\right]^{1/p}. 
 \end{multline}
\end{lem}
\begin{pf}
Let us consider 
\begin{multline}
g(y) = \int f(x,y)\tilde\psi(\sigma_1x-\theta)\,dx = \\
=\frac{1}{\sigma_1^M}\int_{\frac{\theta_1-L}{\sigma_1}}^{\frac{\theta_1+L}{\sigma_1}}
\frac{d^M}{dx^M}f(x,y)
\tilde\Psi_M(\sigma_1 x-\theta_1)\,
dx.
\end{multline}
Then 
\begin{multline}
 <f(x,y),\tilde\psi(\sigma_1 x-\theta_1)\tilde\psi(\sigma_2 y-\theta_2)>=<g(y),\tilde\psi(\sigma_2 y-\theta_2)>=\\
 = \frac{1}{\sigma_1^M\sigma_2^M}
 \int_{\frac{\theta_2-L}{\sigma_2}}^{\frac{\theta_2+L}{\sigma_2}}
 \tilde\Psi_M(\sigma_2 y-\theta_2)
 \frac{d^M}{dy^M}\int_{\frac{\theta_1-L}{\sigma_1}}^{\frac{\theta_1+L}{\sigma_1}}
\frac{d^M}{dx^M}f(x,y)
\tilde\Psi_M(\sigma_1 x-\theta_1)\,
dx\,dy
\end{multline}
Because $\frac{d^{2M}}{dx^Mdy^M}f(x,y)$ and $\tilde\Psi_M$ are absolutely integrable, we can differentiate under the integral sign:
\begin{multline}
 <f(x,y),\tilde\psi(\sigma_1 x-\theta_1)\tilde\psi(\sigma_2 y-\theta_2)>=\\
 =
 \frac{1}{\sigma_1^M\sigma_2^M}
 \int_{\frac{\theta_1-L}{\sigma_1}}^{\frac{\theta_1+L}{\sigma_1}}
\int_{\frac{\theta_2-L}{\sigma_2}}^{\frac{\theta_2+L}{\sigma_2}}
\tilde\Psi_M(\sigma_1 x-\theta_1)
\tilde\Psi_M(\sigma_2 y-\theta_2)
\frac{d^{2M}}{dx^Mdy^M}f(x,y)\, dy\, dx
\end{multline}
By applying the H\"older inequality we conclude the proof.
\end{pf}

\bibliography{wavelet}
\end{document}